\documentclass[10pt,twocolumn,letterpaper]{article}
\usepackage{todonotes}
\usepackage[final]{cvpr}      
\usepackage[export]{adjustbox}
\usepackage[pagebackref,breaklinks,colorlinks]{hyperref}

\def\paperID{*****} 
\def\confName{3DV\xspace}
\def\confYear{2024\xspace}

\title{Geometry Aware Field-to-field Transformations for 3D Semantic Segmentation}

\author{Dominik Hollidt\\
ETH\\
{\tt\small dhollidt@ethz.ch}
\and
Clinton Wang\\
MIT\\
{\tt\small clintonw@csail.mit.edu}
\and
Polina Golland\\
MIT\\
{\tt\small polina@csail.mit.edu}
\and
Marc Pollefeys\\
ETH\\
{\tt\small marc.pollefeys@inf.ethz.ch}
}

\begin{document}
\maketitle
\begin{abstract}
We present a novel approach to perform 3D semantic segmentation solely from 2D supervision by leveraging Neural Radiance Fields (NeRFs). By extracting features along a surface point cloud, we achieve a compact representation of the scene which is sample-efficient and conducive to 3D reasoning. Learning this feature space in an unsupervised manner via masked autoencoding enables few-shot segmentation. Our method is agnostic to the scene parameterization, working on scenes fit with any type of NeRF.
\end{abstract}    
\vspace{-3pt}
\section{Introduction}
\label{sec:intro}

Scene segmentation is a long-standing task in computer vision, playing an important role in robotics, augmented reality (AR), autonomous driving, and many other applications. 
While many existing methods operate in 2D-image space to achieve their goal many applications necessitate 3D segmentations. For example, autonomous driving requires an understanding of what part of the world is occupied \cite{feng_deep_2021} in 3D for accurate path planning, and a 2D label map may be insufficient. Similarly, augmented reality applications may call for dense 3D representations \cite{ming_deep_2021}. Hence, methods that enable dense 3D scene representations and segmentations promise to advance many applications. 

The performance of existing 3D semantic segmentation methods \cite{graham_3d_2018, lawin_deep_2017, qi_pointnet_2017, zhao_point_2021, lai_stratified_2022} lag behind 2D approaches not only due to the difficulty of the task but also due to the lower amount of 3D annotated data \cite{lawin_deep_2017} and higher computational complexity \cite{choy_4d_2019}, e.g., 3D convolutions compared to 2D convolutions.

Some prior work relies on 3D sensors \cite{douillard_segmentation_2011, izadi_kinectfusion_2011}, e.g., LiDAR, or 3D annotation \cite{dai_scannet_2017, caesar_nuscenes_2020, chang_shapenet_2015, armeni_joint_2017, armeni_3d_2016}. However, 3D sensors are less commonly available and expensive. 3D annotations are helpful \cite{graham_3d_2017,qi_pointnet_2017,he_deep_2021} but time-consuming to obtain. Point clouds and structured voxel grids have been used in 3D segmentation \cite{he_deep_2021, graham_submanifold_2017}. Voxel grids can be processed via variations of 3D convolutions \cite{choy_4d_2019, graham_spatially-sparse_2014, li_pointcnn_2018}, while the unstructured point clouds use adaptations of convolutions \cite{thomas_kpconv_2019}, local feature aggregation \cite{qi_pointnet_2017}, or transformer architectures \cite{zhao_point_2021,huang_lcpformer_2023, lai_stratified_2022}. Due to the higher complexity of 3D representations, many approaches are restricted to image space and project their estimates into 3D space, based on estimated depth maps \cite{he_deep_2021}. 

Neural Radiance Fields (NeRFs) have emerged as a powerful method to generate novel views by representing 3D scenes as a neural network \cite{mildenhall_nerf_2020}. NeRFs find many useful applications in graphics \cite{tancik_block-nerf_2022, turki_mega-nerf_2022, poole_dreamfusion_2022, gu_stylenerf_2021}, robotics \cite{fu_panoptic_2022, kundu_panoptic_2022}, and human body representation \cite{zhao_humannerf_2022, peng_neural_2021}. Computer vision tasks like semantic segmentation pose greater challenges for NeRFs compared to explicit representations like point clouds and voxel grids due to the absence of direct field annotations. Instead, discrete labels from either the 2D image space or 3D world must be utilized. Current approaches often require joint training with the NeRF \cite{zhi_-place_2021} or are confined to a specific parameter space, limiting their applicability to NeRFs beyond the one they were trained on. This limitation is noteworthy, given the highly heterogeneous and constantly evolving ecosystem of NeRF datasets.

Similar to \cite{vora_nesf_2021}, we instead adopt a query-based approach that is naturally parameterization-agnostic, so that our model can be trained and deployed on NeRFs of any parameterization. While a parameter-agnostic field-to-field translation has been demonstrated by probing a NeRFs density field with a regular grid \cite{vora_nesf_2021}, we find that sampling the scene's surface geometry leads to a more natural and efficient sampling method that does not waste sampling in empty air regions.

We introduce a simple strategy for semantic segmentation of NeRF and representation learning on neural fields more generally. We render an input NeRF from several coarse views to produce a small point cloud. From that point features are extracted using a stratified transformer model, which can then be used for downstream tasks such as semantic segmentation of novel views or at arbitrary points outside the original point cloud. We investigate masked auto-encoding strategies for pretraining a rich point cloud feature representation and show that using pretrained weights from other tasks can also be helpful in bootstrapping such a network. Our main contributions are as follows: 
\begin{itemize}
    \item We present a simple method for NeRF segmentation that relies only on supervision from 2D views.
    \item We show that the computation can be readily amortized by computing a set of features at a small set of points, which enables for real-time inference and suggests that the feature point cloud can serve as a powerful intermediate representation of the NeRF for downstream tasks.
    \item We show how performing masked auto-encoding of RGB and surface normals from the neural field as a pretraining task can improve the method's performance and data-efficiency on downstream tasks.
\end{itemize}


\vspace{-3pt}
\section{Related Work}\label{sec:related_work}

\subsection{3D Semantic Segmentation}
Semantic segmentation is a fundamental task in computer vision, and many methods have been proposed to tackle it. Most methods work in image space and show impressive results \cite{chen_deeplab_2017, chen_vision_2023, borse_inverseform_2021, wang_internimage_2023}. In the 2D space networks assign a class distribution to each pixel. In the 3D space, point clouds \cite{qi_pointnet_2017, li_pointcnn_2018, engel_point_2021, lai_stratified_2022}, meshes \cite{xu_directionally_2017, hanocka_meshcnn_2019} or voxel grids \cite{choy_4d_2019, graham_spatially-sparse_2014,le_pointgrid_2018, wang_voxsegnet_2018} are used for segmentation. Our method supports simultaneous 2D and 3D semantic segmentation using unstructured point clouds. \\

Neural Semantic Fields (NeSF)  has shown before to enable a 2D and 3D semantic segmentation simultaneously learned purely from 2D supervision via a field-to-field translation from any underlying NeRF that can provide a density field \cite{vora_nesf_2021}. In more detail, NeSF infers a semantic field for a scene by querying a regular grid confined within the known scene boundaries. This grid is then transformed into a feature grid via a shared 3D-UNet. Subsequently, the semantic field can be accessed by interpolating the features of the neighboring grid points and converting them into semantic class probabilities  via a shared MLP. While this method is the first to provide both 3D and 2D semantic segmentation from 2D supervision, it struggles in larger scenes where the scene boundaries are not entirely fixed, due to the cubic complexity of the regular grid and the fact that many grid samples are "empty", i.e. represent air. \\

During our probing step, we obtain the surface geometry represented as a point cloud. Our method is compatible with any differentiable point cloud segmentation network. In our experiments, we utilized PointNet++ \cite{qi_pointnet_2017} and the Stratified Point Transformer (SPT) \cite{lai_stratified_2022}. PointNet++ was one of the first deep-learning approaches to process unordered 3D point clouds. It achieves this by hierarchically applying max pooling operations to local neighborhoods, effectively downsampling the point cloud. To capture both local and global features, multi-scale grouping mechanisms are introduced, enabling feature extraction from neighborhoods at different scales. The SPT makes use of the Transformers \cite{vaswani_attention_2017}. The self-attention memory consumption in the classical transformer architecture scales quadratically in point cloud size and is therefore unsuited for bigger point clouds. The SPT mitigates this problem by processing the point cloud hierarchically in several windows. Each window can capture long-range contexts by sampling points from outside the window via "stratified key-sampling". Initially, points are feature embedded via an adapted convolution KPConv \cite{thomas_kpconv_2019}. The SPT is among the best-performing architectures for 3D semantic segmentation for S3DIS and ScanNetV2.

\subsection{NeRF}
Neural Radiance Fields (NeRFs) are deep neural networks trained on sets of 2D images and their camera pose \cite{mildenhall_nerf_2020} to capture complex 3D scenes. These networks represent scenes as continuous volumetric functions, mapping 3D coordinates and the viewing direction to density and color values. By minimizing differences between rendered and input images, NeRFs produce high-quality novel views with fine details. To render novel views, the 3D field information is integrated into pixel values by shooting rays into the scene, represented as $\mathbf{r}(t) = o + td$, where $o$ is the origin and $d$ is the direction of the ray, with $t \in \mathbb{R}^+$. The volumetric rendering equation is then used to obtain the ray's color $C(r)$. This involves acquiring the density $\sigma(x)$, which represents the differential probability of a ray terminating at location $x$ \cite{mildenhall_nerf_2020}, as well as the colors $\mathbf{c}(x, d)$ of samples along the ray:
\begin{equation}
    \begin{gathered}
          C(\mathbf{r})=\int_{t_n}^{t_f} T(t) \sigma(\mathbf{r}(t)) \mathbf{c}(\mathbf{r}(t), \mathbf{d}) d t, \\
          \text { where } T(t)=\exp\left(-\int_{t_n}^t\sigma(\mathbf{r}(s))ds\right).
    \end{gathered}
    \label{eq:volume_rendering}
\end{equation}
Here, $T(t)$ represents the accumulated transmittance along the ray \cite{mildenhall_nerf_2020}.  NeRFs can provide additional outputs beyond density and RGB values. For instance, they may include normals $n$ \cite{verbin_ref-nerf_2022} or semantics \cite{zhi_-place_2021} in the form of $(x, y, z, \phi, \psi) \rightarrow (r, g, b, \sigma, n_x, n_y, n_z, s_1, s_2, \dots, s_i)$, where $s_i$ represents the probability of class $i$. In such cases, the semantic label distribution is denoted by $L(r)$.
\begin{equation}\label{eq:semantic_volumetric_rendering}
    L(\mathbf{r})=\int_{t_n}^{t_f} T(t) \sigma(\mathbf{r}(t)) \mathbf{s}(\mathbf{r}(t), \mathbf{d}) d t.
\end{equation}

Several NeRF approaches \cite{muller_instant_2022, liu_neural_2021, xu_grid-guided_2023} aim to improve inference time by encoding the scene in feature data structures, particularly grid and hashmap-based methods. In contrast, PointNeRF \cite{xu_point-nerf_2022} employs point clouds to, approximate the scene geometry where each point has a learnable feature vector. At test time, a small MLP fuses local neighbors to provide the final radiance and density. We draw inspiration from this approach for our architecture.

\subsection{Pretraining}
Many successful pretraining tasks have been proposed for point cloud pretraining \cite{zhang_self-supervised_2021, xie_pointcontrast_2020, wang_unsupervised_pretraining_2021}. 
Masked auto-encoding, originally used for pretraining vision transformers on 2D images \cite{he_masked_2021}, masks a portion of the input image and passes it through an encoder, and a smaller decoder is trained to fill in the masked pixels. This technique has been adapted for point clouds in various forms \cite{avidan_masked_2022, zhang_masked_2022} and used in this paper.
\vspace{-4pt}
\section{Method}\label{sec:method}
Our method aims to generalize over many scenes while working with any NeRF parameterization. To this end, we perform efficient surface geometry probing of the scene and use 3D point cloud segmentation. The method can integrate 3D labels into 2D semantic maps using volumetric rendering \cref{eq:volume_rendering}, to support 3D semantic segmentation of the point cloud from pure 2D supervision. Our training proceeds in two steps: First we fit NeRFs to each scene in the training dataset (\cref{sec:nerf_training}). Secondly, we train a shared transformation model that predicts semantic labels for a novel view via surface geometric aware sampling and a shared point cloud segmentation model, described in \cref{sec:transformation_network}. 

Further, we show how to amortize the computation with a point cloud feature representation so that segmentations at new points can be queried in real-time (\cref{sec:field2field}). Lastly, we utilize additional information encoded in the NeRF in a pretraining step for our method to achieve better accuracy in data-scarce scenarios (\cref{sec:pretraining}).

\subsection{NeRF Training}\label{sec:nerf_training}
Our scene transformation works with any NeRF parameterization. Here we employ the proposal network used in most NeRF variants \cite{barron_mip-nerf_2021, mildenhall_nerf_2020} to obtain a surface point cloud more efficiently.

We require a collection of $\mathbf{s}$ scenes, each stored as a collection of RGB images $\{\mathcal{C}^{gt}_{s,c} \in [0,1]^{H\times W \times 3}\}$ paired with their camera parameters $\mathcal{P} = \{p_{s,c} \in \mathbb{R}^{\Gamma}\}$, semantic maps $\{S^{gt}_{s,c} \in \mathbb{Z}^{H\times W}\}$ and optional depth maps $\{\mathcal{D}^{gt}_{s,c} \in R^{H\times W}\}$, with $s$ representing the scene and $c$ the camera index. It is not necessary for RGB images and semantic maps to have a one-to-one correspondence. We denote the set of all rays corresponding to a pixel of an image by $\mathcal{R}(p_{s,c})$. If point $x$ belongs to a ray $\mathbf{r} \in \mathcal{R}(p_{s,c})$ we write $x \in \mathbf{r}$.

We fit each scene with Nerfacto \cite{tancik_nerfstudio_2023}, a fast, robust variant of Instant NGP \cite{muller_instant_2022} that uses a multi-resolution hash-based feature encoding, and incorporates other techniques such as proposal sampling as in Mip-NeRF-360 \cite{barron_mip-nerf_2022}. It is trained via volumetric rendering (\cref{eq:volume_rendering}) and uses a combination the photometric reconstruction loss \cite{tancik_nerfstudio_2023, mildenhall_nerf_2020}, the proposal loss and distortion loss as in Mip-NeRF-360 \cite{barron_mip-nerf_2022}. \\

During pretraining, we utilize the normals of the NeRF, which can be extracted analytically by differentiating the density with respect to the position:
\begin{equation}
    \mathbf{n}(x) = - \frac{\delta \sigma(x)}{\delta x}
\end{equation}
Since the normals often exhibit considerable noise, Nerfacto incorporates an additional head to predict normals alongside RGB and density, resulting in smoother normal predictions. The prediction process involves the use of a combination of the dot product loss and normalizing loss, as demonstrated in Ref-NeRF \cite{verbin_ref-nerf_2022}. We observed that Nerfacto encounters difficulties with floaters, and to mitigate this effect, we introduce minimal depth supervision \cite{deng_depth-supervised_2022}.

\subsection{Field Transformation}
The transformation network probes the NeRF and reasons about the 3D semantic classes within the scene. This process is divided into two steps: 1. probing the NeRF to obtain a surface represented as a point cloud, 2. reasoning about the 3D semantic classes. We also propose an intermediate feature representation of the point cloud that can be trained in an unsupervised manner.

\begin{figure*}[!htbp]
  \centering
    \includegraphics[width=\linewidth]{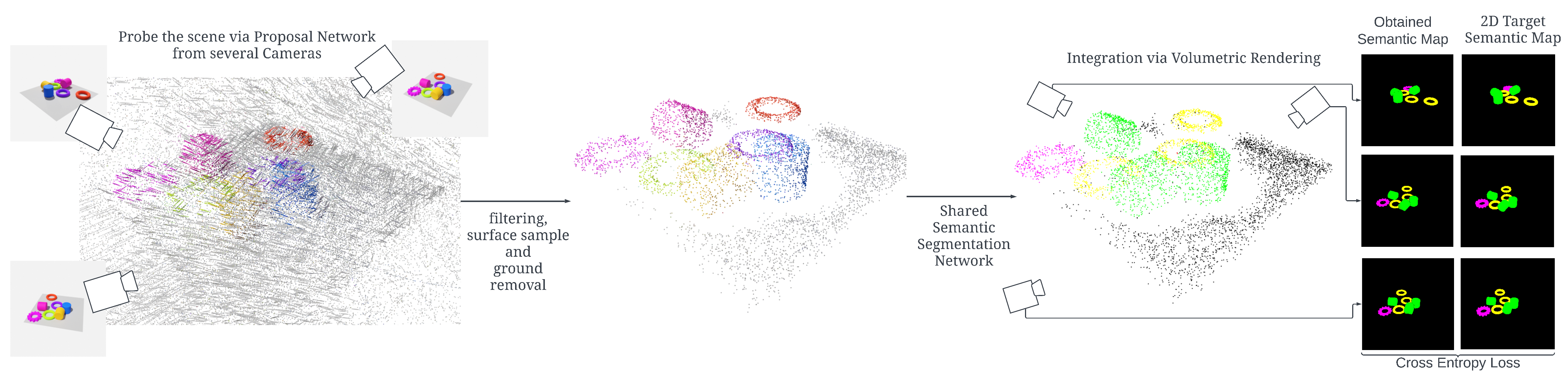}
  \caption{A visualization of our method. (1) We shoot rays from several poses into the scene and sample points according to the proposal network of the NeRF. (2) We obtain a point cloud capturing the surface geometry of the scene by applying scene boundary filtering, density filtering, surface sampling and ground removal. (3) The point cloud is segmented via a semantic segmentation network and integrated via volumetric rendering (4) The network gets trained with the 2D semantic maps and cross entropy loss.}
  \label{fig:nesf}
  \vspace{-5mm}
\end{figure*}

\subsubsection{Probing}
To obtain a concise yet geometrically meaningful scene representation, we probe the scene's surface geometry. To connect the 3D label to the 2D semantic map via volumetric rendering in \cref{eq:semantic_volumetric_rendering}, sampling is restricted to points on rays, denoted as $\hat{X}=\{x | \exists \mathbf{r} \in \mathcal{R}_s: x\in \mathbf{r} \}$. Therefore, we sample pixels from the scene, obtaining corresponding rays, and subsequently, we sample $k$ points along each ray weighted by density via the NeRF's proposal network \cite{barron_mip-nerf_2022}, forming $X_{prop} \subset \hat{X}$. To reduce the number of samples we filter the point cloud to represent only the surface scene geometry via four filtering techniques: point scene-bound filtering, density filtering, surface sampling, and ground plane removal. Initially, we remove points that lie outside a rough estimate of the known scene bounds. Subsequently, we discard points below a certain density threshold, as they typically correspond to empty regions. To identify the surface point along the ray let $w_1, w_2, \ldots, w_n$ be a list of weights, where $n$ represents the total number of samples amount the ray. We define the cumulative weight $A_i$ as the sum of the first $i$ weights:
\[ A_i = \sum_{k=1}^{i} w_k. \]
To find the first weight that has more than $\eta$ of the accumulated weights, we identify the smallest index $j$ such that:
\vspace{-1em}
\[ \frac{A_j}{\sum_{k=1}^{n} w_k} > \eta. \]
\vspace{-0.5em}
Specifically, we find the first weight $w_j$ that satisfies
\[ A_j > \eta \times \left(\sum_{k=1}^{n} w_k\right). \]

Many points belong to the floor of the scene, yet we do not want to waste resources on semantically labeling the floor. We filter out points from the floor via RANSAC with 100 fitted planes. Finally, we remove all points which are close to and below the best-fitting plane, i.e. $sd(x, plane) < d_{plane}$. Overall, the combination of these filtering steps produces a significant reduction of the point cloud size (up to 100x) while preserving relevant surface geometry. All filter steps yield a point cloud $\mathcal{X} = filter(X_{prop})$ and do not interfere with volumetric rendering. 

\subsubsection{Transformation Network}\label{sec:transformation_network}
The transformation network takes the probed point cloud and its features as input and generates an intermediate point cloud with associated features. Each probed point's coordinates are concatenated with features such as sinusoidal position encoding (following \cite{mildenhall_nerf_2020}), spherical harmonics encoding the viewing direction, RGB values, and min-max normalized density. The sampled points, along with their features $\mathcal{F}_{\mathcal{X}}$ are transformed into an intermediate feature point cloud, which can be reduced to semantic classes via a shared MLP $\mathcal{S} = MLP_{sem}(\mathcal{T}(\mathcal{X}, \mathcal{F}_{\mathcal{X}}))$. We use random rotations along the z-axis as data augmentation and normalize the point cloud to be in $[0,1]^3$ via translation and uniform scaling across all dimensions to not distort it. Using volumetric rendering, as in \cref{eq:semantic_volumetric_rendering}, we obtain a distribution over class per ray and train the network via the cross entropy loss with the 2D semantic map as supervision. To encourage locally smooth predictions we use a proximity loss similar to NeSF \cite{vora_nesf_2021}. This step adds jitter to the original point cloud and nudges the model to predict the same features on the jittered point cloud as for the original point cloud via MSE loss:
\begin{equation}
    \mathcal{L}_{proximity} = |\mathcal{T}_{logits}(\mathcal{X}, \mathcal{F_{\mathcal{X}}}) - \mathcal{T}_{logits}(\mathcal{X}_{jitter}, \mathcal{F}_{\mathcal{X}_{jitter}})|^2
\end{equation}
The jitter uses Gaussian noise with $\sigma=0.003$. The proximity loss is weighted with $\lambda_{proximity}=0.01$. For $\mathcal{T}$ we use a transformer architecture, PointNet++ \cite{qi_pointnet_2017}, and the Stratified Point Transformer \cite{lai_stratified_2022}.

\subsection{Field-to-Field Transformation}\label{sec:field2field}
\begin{figure}[!htbp]
    \centering
    \includegraphics[width=0.5\textwidth]{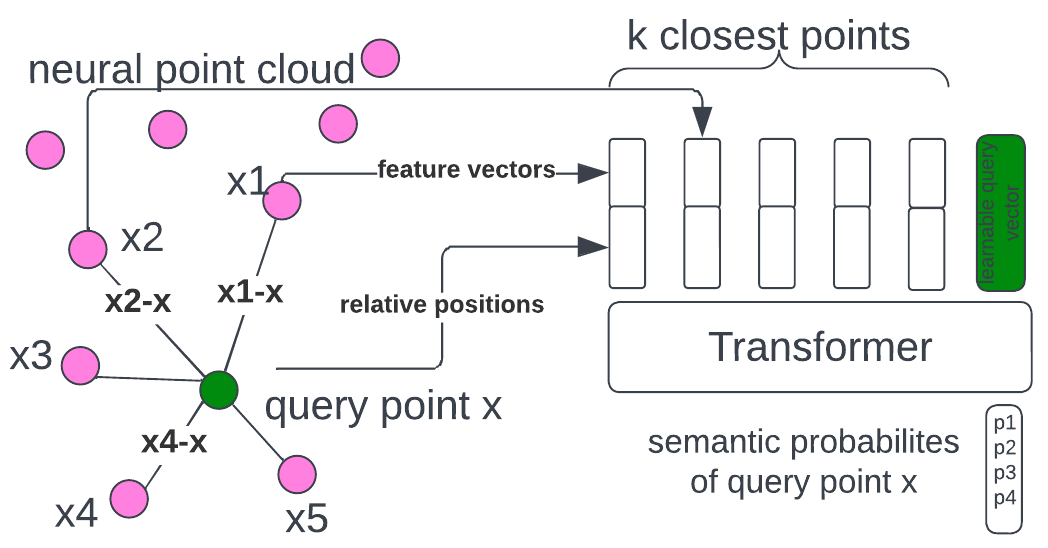}
    \caption{The Field Head takes the $k$ closest points and their features together with the relative position to provide fast inference of their class probabilities.}
    \label{fig:field_head}
    \vspace{-3mm}
\end{figure}

We train the transformation network to produce a set of features along a small point cloud, and append a field head that learns to predict the segmentation label of any query point given this feature point cloud. This enables our model to compute the features once for a scene and then perform segmentation in real time for novel views. Inspired by Point-NeRF \cite{xu_point-nerf_2023} we treat our transformed point cloud as a neural point cloud and learn a feature vector per point via the transformation network. For one query point the k nearest neighbor's features concatenated with their relative position to the query point $r_i$ are passed into a shared transformer $FT$. Additionally, we prepend a learnable query token $\mathbf{Q}$ to the sequence of $k$ vectors (\cref{fig:field_head}). 
\begin{equation}
    f_{query} = FT( \mathbf{Q}, (f_1, r_1), (f_2, r_2) , ..., (f_k, r_k))
\end{equation}
Finally, the semantic labels at the query point are estimated by passing query tokens feature encoding $f_{query}$ through a shared MLP:
\begin{equation}
    \mathbf{s}_{query} = MLP_{field}(f_{query}).
\end{equation}
In order to learn our semantic field we proceed as before. First, we sample a neural point cloud and extract its features with $\mathcal{T}$. Secondly, we sample our query point cloud and obtain the semantic probabilities per point via $FT$ and $MLP_{field}$. All networks can be trained end-to-end via backpropagation of the cross entropy loss through the volumetric rendering equation \cref{eq:volume_rendering}.

\subsection{Pretraining}\label{sec:pretraining}
Pretraining for point clouds has already proven successful as described in \cref{sec:related_work}. Therefore, applying existing pretraining techniques for point cloud pretraining is straightforward. We also investigate whether the additional available information provided by the NeRF, i.e., RGB, density, and normals, can pose a challenging pretraining task to benefit downstream semantic segmentation. \\

\begin{figure*}[!htbp]
    \begin{minipage}{.48\textwidth}
    \includegraphics[width=\textwidth]{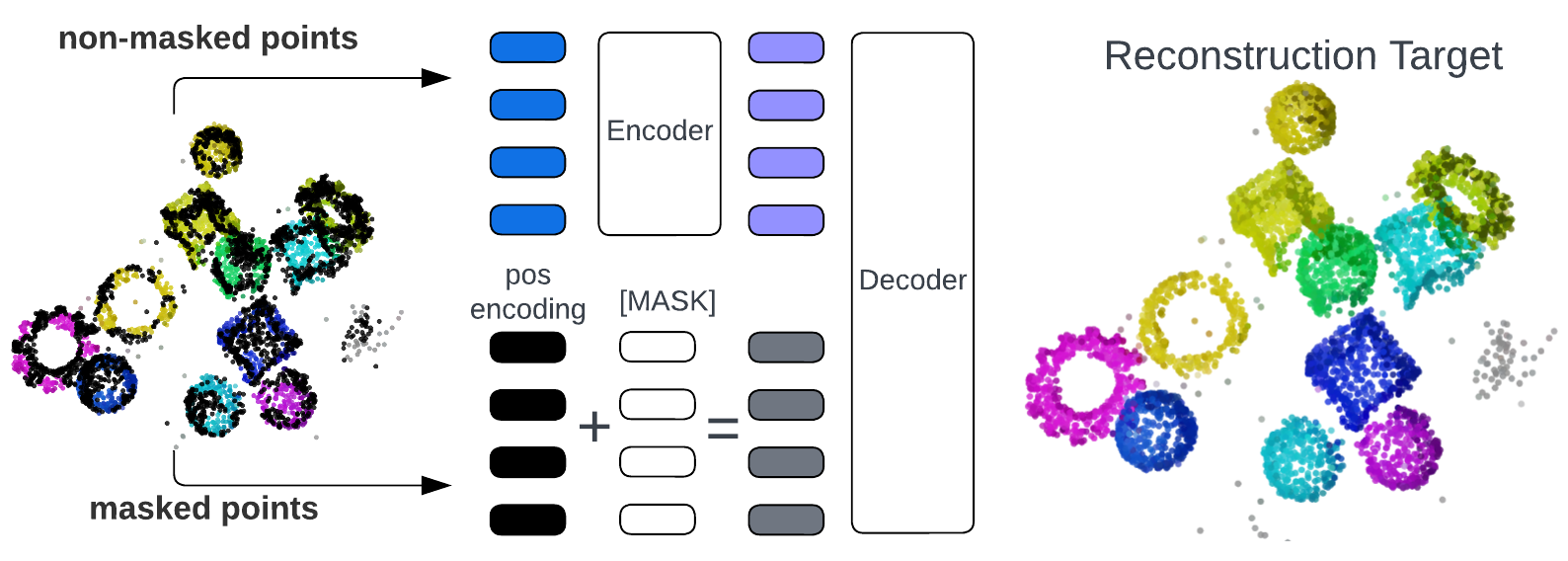}
    \caption{In the pretraining masked autoencoding stage part of the point cloud gets masked and properties of it have to be recovered from features extracted from the non-masked point cloud and just the position of the masked point cloud.}
    \label{fig:masked_auto_encoding}
    \end{minipage}
    \hfill
    \begin{minipage}{.50\textwidth}
        \begin{tabular}{l|c|c|c}
             & KLEVR & ToyBox5 & Kubasic \\
            \hline \# scenes & $100 / 20$ & $500 / 25$ & $80 / 20$ \\
            \# cameras/scene & $270 / 30$ & $270 / 30$ & $230 / 26$\\
            frame resolution & $256 \times 256$ & $256 \times 256$ & $256 \times 256$ \\
            \# objects/scene & $4-12$ & $4-12$ & $3-7$\\
            \# object instances & 5 & 25,905 & 11\\
            \# background instances & 1 & 383 & 100\\
        \end{tabular}
    \caption{Datasets are split into train and test scenes. Scenes are split into train and evaluation images. }
    \label{tab:data}
    \end{minipage}
    \vspace{-5mm}
\end{figure*}

NeRFs provide RGB information for each point in 3D. Therefore, we use masked auto encoding to recover their RGB values (\cref{fig:masked_auto_encoding}). The decoder is smaller in size and only has 40\% of the parameters of the encoder. We investigate random masking, patch masking and patch masking with furthest point sampling. For patch masking $k$ key points $K = \{x | \text{$x$ is key point}\}, \text{such that } |K| = k$ are selected. In the second step, we compute the distance from any point to the closest key point:
\begin{equation*}
    dist(x, K) = \min\limits_{x'\in K} {|x-x'|}
\end{equation*}
Then we select the closest $p$ fraction of the points as being masked. In patch masking with furthest point sampling, we select key points using furthest point sampling, leading to more evenly scattered patches. We train the RGB pretraining objective with mean absolute error. \\

Pretraining with RGB alone might not encourage the model to learn local shapes effectively, but rather focus on interpolating colors. To address this, we introduce predicting normals as a second pretraining objective. Learning normals for each point prompts the network to understand neighborhoods and their shapes. We utilize the dot product loss for the normal pretraining objective.
\vspace{-3pt}
\section{Experiments}\label{sec:experiments}
We evaluate our method on two datasets, KLEVR and ToyBox5 \cite{vora_nesf_2021}. Additionally, we generated our own dataset to evaluate the pretraining of normals with Kubric \cite{greff_kubric_2022}. Our datasets require a larger number of scenes to evaluate the generalizability of our method. Each scene provides camera poses together with the corresponding RGB image, semantic and depth map.

The KLEVR dataset epitomizes a rudimentary, synthetically-derived proof-of-concept exemplar comprising five object categories, such as cubes and cylinders, arbitrarily oriented, colored, and scaled on a consistently gray plane. Conversely, the ToyBox5 dataset engages with a more sophisticated assortment of object categories, encompassing chairs, tables, and airplanes. The instances of these objects are selected from the ShapeNet database \cite{chang_shapenet_2015}, and placed with a random scale on the floor in front of realistic High Dynamic Range Imaging (HDRI) backgrounds. Kubasic is a slight increase in difficulty over KLEVR as it includes HDRI backgrounds and adds 6 more object classes. Additionally, the objects are placed anywhere within the scene. For segmentation purposes, the background and floor components are incorporated as an extra class. 

In our methodology, a defined quantity of scenes is allocated for training purposes, with a separate, non-overlapping segment designated as test scenes. Within each training scene, a significant proportion of cameras is employed for training, while a distinct subset serves to monitor and validate the progression of training. Ten randomly selected views in test scenes are used for evaluation.

We train each Nerfacto for 10K steps using an Adam optimizer with a learning rate of 0.01, and 4096 rays per step. We provide use $\lambda_{prop}=1$, $\lambda_{dist}=0.002$, $\lambda_{normal}=0.01$, $\lambda_{normal\_reg}=0.0001$ and $\lambda_{depth}=0.01$. Training takes 12 minutes on a single QRTX5000 GPU. The NeRFs achieve good visual fidelity with a SSIM of 0.92-0.96, PSNR of 32-40, and depth-MSE typically smaller than 0.002 on both KLEVR and ToyBox5.

The following sections provide results on the best models, the impact of the field-to-field head, and the benefits of pretraining. Additionally, we provide many ablation studies in the supplementary materials to justify our design choices. 

\vspace{-2mm}
\subsection{Evaluation}
The evaluation reports the 2D-mIoU over the whole test set, i.e., accumulating the statistics over the test set and then computing the mIoU. We compare our results to those reported in \cite{vora_nesf_2021}, i.e., NeSF and DeepLab v3 \cite{chen_deeplab_2017} - a 2D segmentation model.

For our transformation model, we compare a custom transformer model, PointNet++ \cite{qi_pointnet_2017}, and the stratified point transformer \cite{lai_stratified_2022}. Our custom transformer architecture dissects the point cloud into multiple vertical pillars, each containing 1,536 points. Every pillar is then processed by a standard transformer encoder equipped with six layers and eight heads. We compare to PointNet++ \cite{xu_pytorch_2023} with a radius scale of 0.2. During the training phase of our stratified point transformer, we utilize 65,536 rays per scene, adhering to the original implementation. We reduce the grid size to 0.005 and the point quantification size to 0.0001. We set the surface threshold ($\mathcal{W}$) to 0.2 for KLEVR scenes and at 0.5 for ToyBox. The Field Head's transformer has 2 layers and 4 heads with a feature dimension of 64. From our empirical observations, a ground tolerance ($d_{plane}$) of 0.0 to 0.005 presents an optimal balance between point cloud size and point removal. We utilize the Adam optimizer with exponential learning rate decay from $5\cdot10^{-3}$ to $5\cdot 10^{-4}$ over 80K steps. Moreover, for all transformer blocks within the architecture, the learning rates are reduced by a factor of 0.1. In the supplementary material, we show how the SPT outperformed other point cloud segmentation network architectures.

The models are trained until they converge, which typically occurs after 20K to 80K steps on KLEVR and between 100K to 200K steps on ToyBox5. The training process spans a duration of one to four days.

\subsection{Quantitative Comparison}
\begin{figure}[!htbp]
    \centering
    \includegraphics[width=0.5\textwidth]{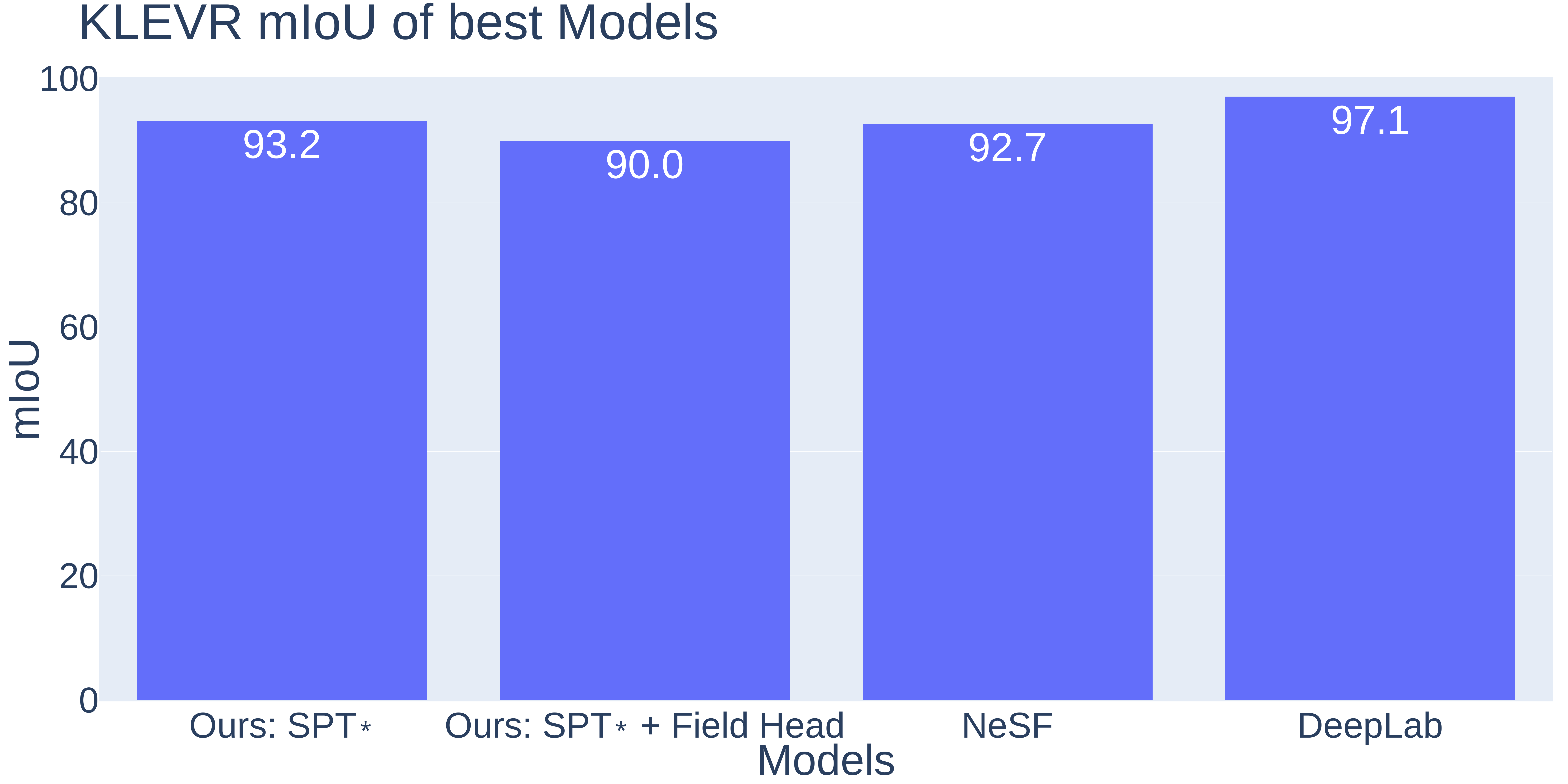}
    \caption{The mIoU of the best-performing methods on KLEVR. Our method is beating NeSF by a small margin. * indicates the use of S3DIS pretrained weights.}
    \label{fig:klevr_best}
    \vspace{-5mm}
\end{figure}

The first set of experiments compares the best-performing models on both KLEVR and the more realistic ToyBox5 dataset. \cref{fig:klevr_best} shows that our method with SPT beats NeSF by a small margin, but adding the field-to-field head reduces the performance for the KLEVR dataset by 3 mIoU. On ToyBox5 our method using the SPT achieves higher accuracy than NeSF and DeepLab. Here, our best model uses the pretrained weights from S3DIS which increases our performance marginally when using all training data (more about pretraining in \cref{sec:pretraining}). Our network is sample efficient. Compared to NeSF, which requires grids of up to $80\times 80\times 80=512K$ points, we only use 5,800 points (88$\times$ less) on KLEVR and 18K points (28$\times$ less) on ToyBox5. This sample efficiency results in significantly lower memory consumption -- NeSF's 3D UNet requires 36GB of allocated VRAM, while our method uses only 10GB on average. The Field Head scales linearly in memory consumption with the number of query points, leading to 22GB usage on average.

\subsection{Qualitative Comparison}
\begin{figure*}[!htbp]
    \centering
    \begin{tabular}{cccccc}
        \begin{subfigure}[b]{0.12\textwidth}
            \centering
            \includegraphics[width=\textwidth]{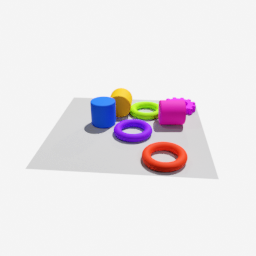}
        \end{subfigure} &
        \begin{subfigure}[b]{0.12\textwidth}
            \centering
            \includegraphics[width=\textwidth]{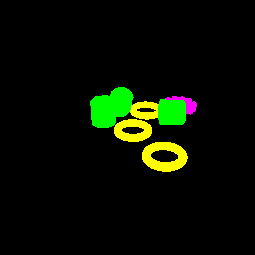}
        \end{subfigure} &
        \begin{subfigure}[b]{0.12\textwidth}
            \centering
            \includegraphics[width=\textwidth]{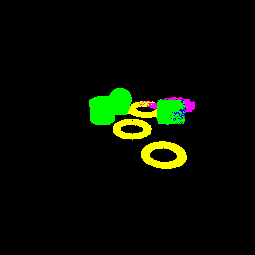}
        \end{subfigure} &
        \begin{subfigure}[b]{0.12\textwidth}
            \centering
            \includegraphics[width=\textwidth]{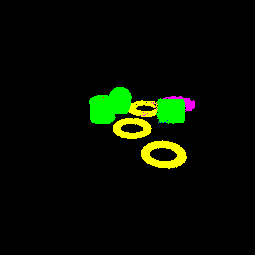}
        \end{subfigure} &
        \begin{subfigure}[b]{0.12\textwidth}
            \centering
            \includegraphics[width=\textwidth]{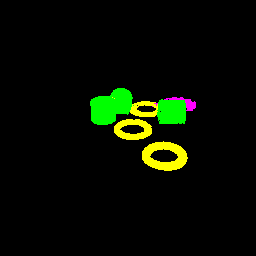}
        \end{subfigure} &
        \begin{subfigure}[b]{0.12\textwidth}
            \centering
            \includegraphics[width=\textwidth]{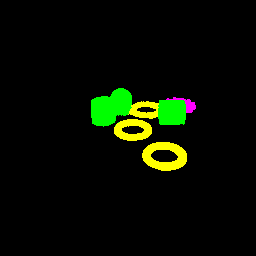}
        \end{subfigure} \\
        \begin{subfigure}[b]{0.12\textwidth}
            \centering
            \includegraphics[width=\textwidth]{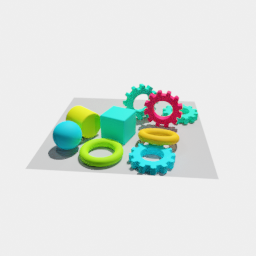}
        \end{subfigure} &
        \begin{subfigure}[b]{0.12\textwidth}
            \centering
            \includegraphics[width=\textwidth]{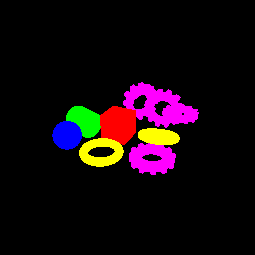}
        \end{subfigure} &
        \begin{subfigure}[b]{0.12\textwidth}
            \centering
            \includegraphics[width=\textwidth]{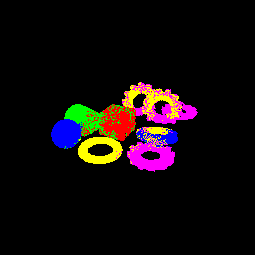}
        \end{subfigure} &
        \begin{subfigure}[b]{0.12\textwidth}
            \centering
            \includegraphics[width=\textwidth]{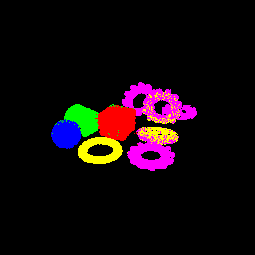}
        \end{subfigure} &
        \begin{subfigure}[b]{0.12\textwidth}
            \centering
            \includegraphics[width=\textwidth]{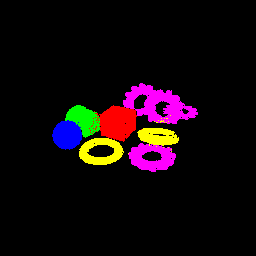}
        \end{subfigure} &
        \begin{subfigure}[b]{0.12\textwidth}
            \centering
            \includegraphics[width=\textwidth]{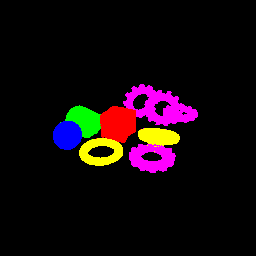}
        \end{subfigure} \\
    \end{tabular}
    \centering
    \begin{tabular}{cccccc}
        \begin{subfigure}[b]{0.12\textwidth}
            \centering
            \includegraphics[width=\textwidth]{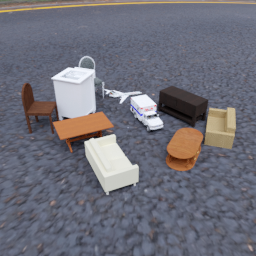}
        \end{subfigure} &
        \begin{subfigure}[b]{0.12\textwidth}
            \centering
            \includegraphics[width=\textwidth]{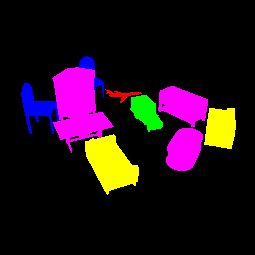}
        \end{subfigure} &
        \begin{subfigure}[b]{0.12\textwidth}
            \centering
            \includegraphics[width=\textwidth]{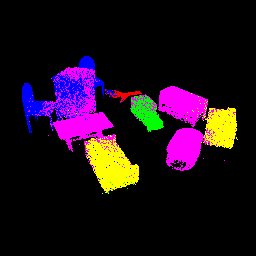}
        \end{subfigure} &
        \begin{subfigure}[b]{0.12\textwidth}
            \centering
            \includegraphics[width=\textwidth]{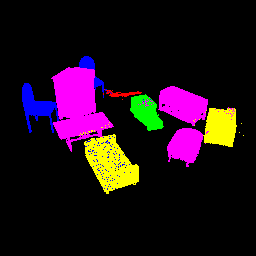}
        \end{subfigure} &
        \begin{subfigure}[b]{0.12\textwidth}
            \centering
            \includegraphics[width=\textwidth]{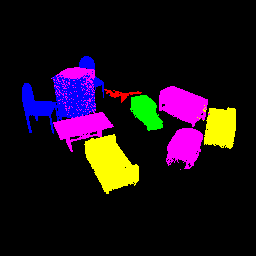}
        \end{subfigure} &
        \begin{subfigure}[b]{0.12\textwidth}
            \centering
            \includegraphics[width=\textwidth]{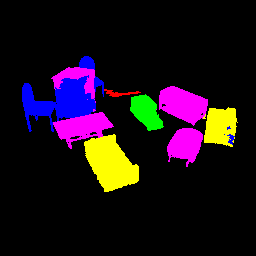}
        \end{subfigure} \\
        \begin{subfigure}[t]{0.12\textwidth}
            \centering
            \includegraphics[width=\textwidth]{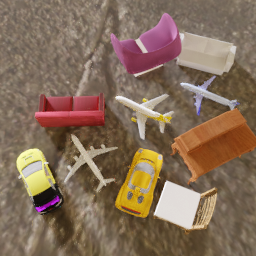}
            \caption{RGB}
        \end{subfigure} &
        \begin{subfigure}[t]{0.12\textwidth}
            \centering
            \includegraphics[width=\textwidth]{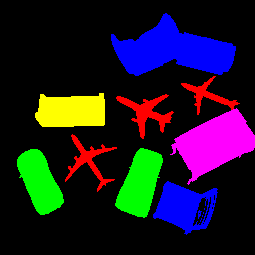}
            \caption{Ground Truth}
        \end{subfigure} &
        \begin{subfigure}[t]{0.12\textwidth}
            \centering
            \includegraphics[width=\textwidth]{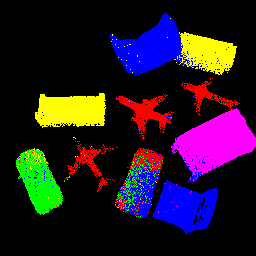}
            \caption{PointNet++}
        \end{subfigure} &
        \begin{subfigure}[t]{0.12\textwidth}
            \centering
            \includegraphics[width=\textwidth]{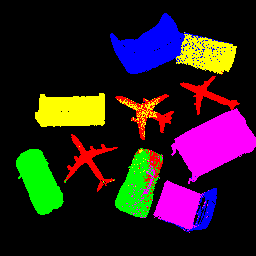}
            \caption{Plain Transformer}
        \end{subfigure} &
        \begin{subfigure}[t]{0.12\textwidth}
            \centering
            \includegraphics[width=\textwidth]{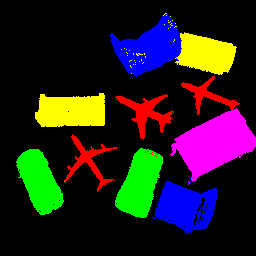}
            \caption{SPT}
        \end{subfigure} &
        \begin{subfigure}[t]{0.12\textwidth}
            \centering
            \includegraphics[width=\textwidth]{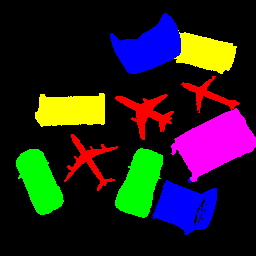}
            \caption{SPT + Field Head}
        \end{subfigure} \\
    \end{tabular}
    \caption{Qualitative comparison on KLEVR (top 2 rows) and ToyBox5 (bottom 2 rows)}
    \label{fig:qualitative}
\end{figure*}

\cref{fig:qualitative} shows that PointNet++ and the custom transformer architecture are less consistent in predicting the class label consistently within an object, and perform worse overall. The SPT performs better than these methods but still sometimes predicts multiple classes within the same object. We notice most misclassifications occur between the chair and sofa class. This becomes evident in the last row of \cref{fig:qualitative}, where a sofa and chair that appear almost identical are labeled differently. The Field Head yields smoother predictions and tends to commit to a single label, yielding more natural-looking results but loses some fine-detailed structures by filling in the gaps. The surface sampling regime provides crisp object boundaries and even captures fine details like the chair legs with significantly fewer points than NeSF. 

\begin{figure}[!htbp]
    \centering
    \includegraphics[width=0.5\textwidth]{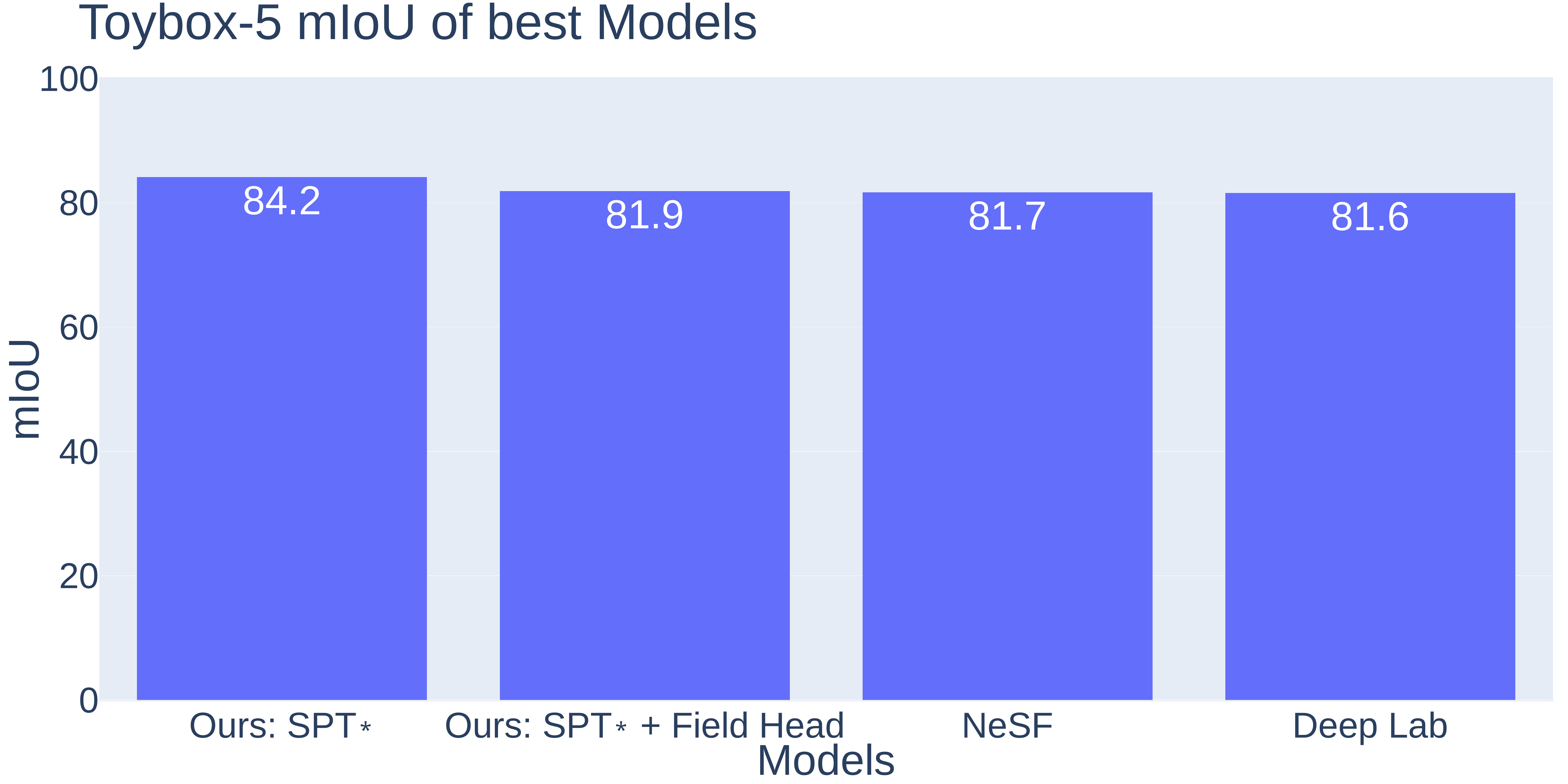}
    \caption{The mIoU of the best-performing methods on ToyBox5. Our method outperforms NeSF and Deep Lab in accuracy. * indicates the use of S3DIS pretrained weights.}
    \label{fig:toybox_best}
    \vspace{-1mm}
\end{figure}

\begin{figure}[!htbp]
    \centering
    \includegraphics[width=0.5\textwidth]{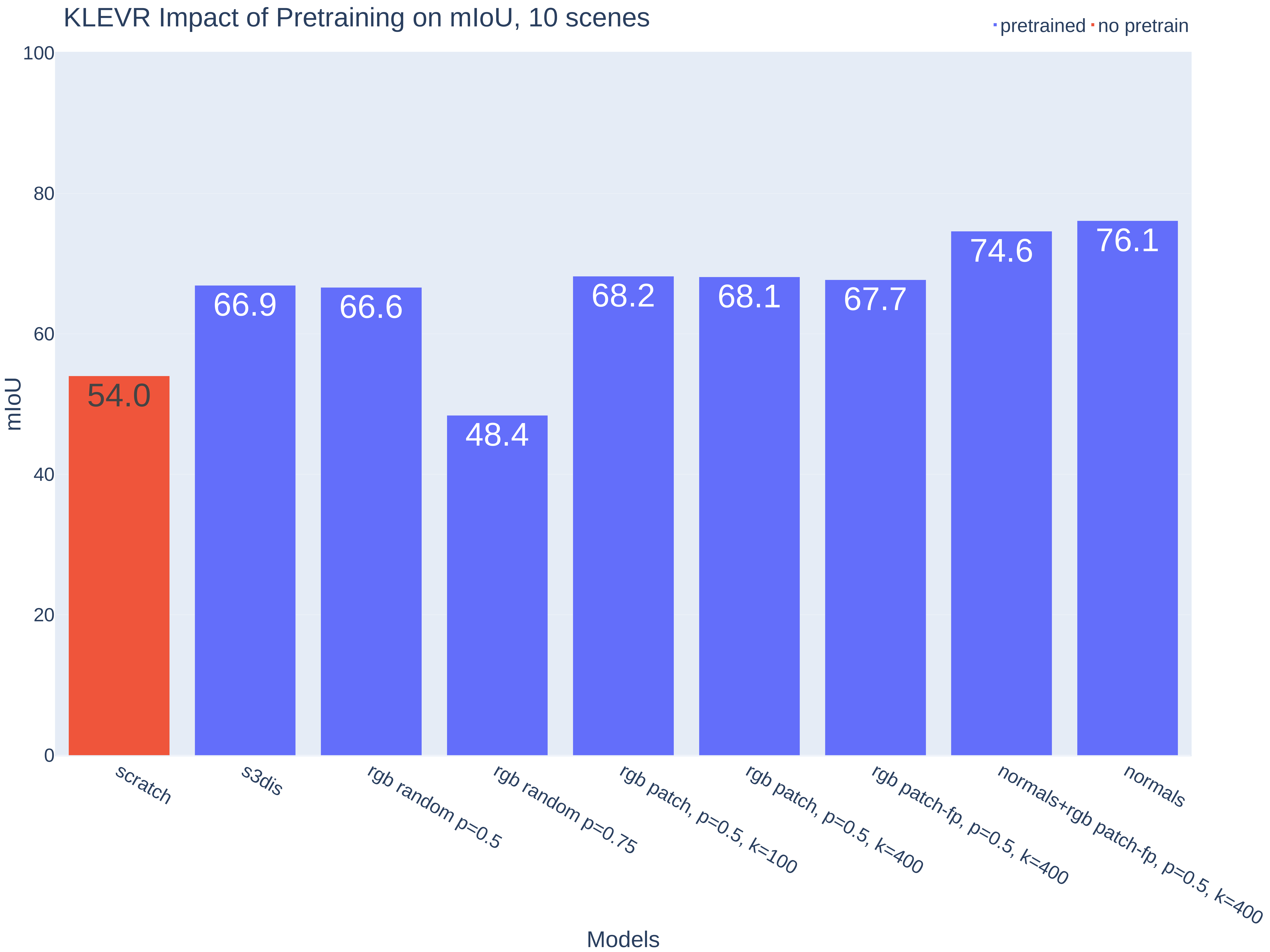}
    \caption{Pretraining has a positive effect for the SPT on KLEVR where only 10\% of the data is used. Normal pretraining particularly boosts the accuracy.}
    \label{fig:klevr_pretraining}
\end{figure}

\subsection{Field-to-Field Head}
The Field Head's main goal is to allow for an accelerated field-to-field translation by precomputing a neural point cloud and using the Field Head to interpolate for a query point. The Field Head offers a significant speed up during inference time, especially for single point queries, as it does not need to run a high complexity network nor does it have to add a point cloud as context when querying just a single point (\cref{tab:inference_speed}). The inference time includes finding the $k$ closest neighbors. At the same time, it decreases the accuracy slightly (\cref{fig:klevr_best,fig:toybox_best}). We observe that the Field Head tends to generate smoother and more natural-looking predictions, as evident in \cref{fig:qualitative}. However, this smoothness occasionally results in larger patches of the wrong label, as seen in the rightmost panel in the second-to-last row of \cref{fig:qualitative}. In many instances, the Field Head produces desirable results by alleviating local grainy ambiguity within the segmentation at the cost of losing some fine-detailed structures whose gaps get filled in by the surrounding label, e.g. the gaps between the gears teeth. 

\begin{table}[!htbp]
	\centering
	\begin{tabular}{c|c}
		Method (query points) & Inference Speed \\
		\hline
		SPT (16,384) & 196ms \\
		Field Head (16,384) & 109ms \\
        SPT (1) & 196ms \\
		Field Head (1) & 2ms \\
	\end{tabular}
	\caption{The Field Head speeds up inference significantly by using a cached neural point cloud (16,384 points). The SPT always needs an even point cloud of the scene as context, hence inference speed of a single point does not increase. }
	\label{tab:inference_speed}
\end{table}

\subsection{Pretraining}
To evaluate the efficacy of pretraining in data-scarce scenarios, we conduct three distinct experiments. The pretraining phase is universally applied to all training scenes. In the KLEVR experiment, fine-tuning training data is restricted to 10 scenes. For ToyBox5, two separate experiments are performed with 100 scenes and either 10 or 270 images for fine-tuning.  We compare the normal and RGB pretraining against a semantic segmentation pretrained model on S3DIS, where we reuse 84\% of the weights, due to a difference in feature dimension. 
As illustrated in \cref{fig:klevr_pretraining}, our pretraining approach proves advantageous in extreme data-scarce situations on KLEVR. All RGB pretrained methods yield approximately equivalent performance, regardless of masking ratio or patching method, although random RGB masking with $p=0.75$ was identified as an outlier. Conversely, normal pretraining proves more beneficial than utilizing S3DIS pretrained weights; the addition of RGB to the pretraining objective leads to a decrease in performance. We hypothesize that normal pretraining endows the model with an understanding of the geometric properties of the objects, which aligns more closely with the semantic segmentation fine-tuning task than the class domain shift inherent in the S3DIS pretraining stage. \\
In the ToyBox5 experiments, RGB pretraining is found to adversely affect weight initialization, leading to significantly reduced accuracy. This discrepancy is attributable to the single-color-per-object nature of KLEVR objects, which enables a coherent understanding of object boundaries that aligns with semantic segmentation. This alignment is absent in ToyBox5. Nonetheless, pretraining with normals still surpasses the randomly initialized baseline by 0.8-0.9\% and performs comparably to, or only 1\% less effectively than, the S3DIS pretrained weights.

\begin{figure}[!htbp]
    \centering
    \includegraphics[width=0.5\textwidth]{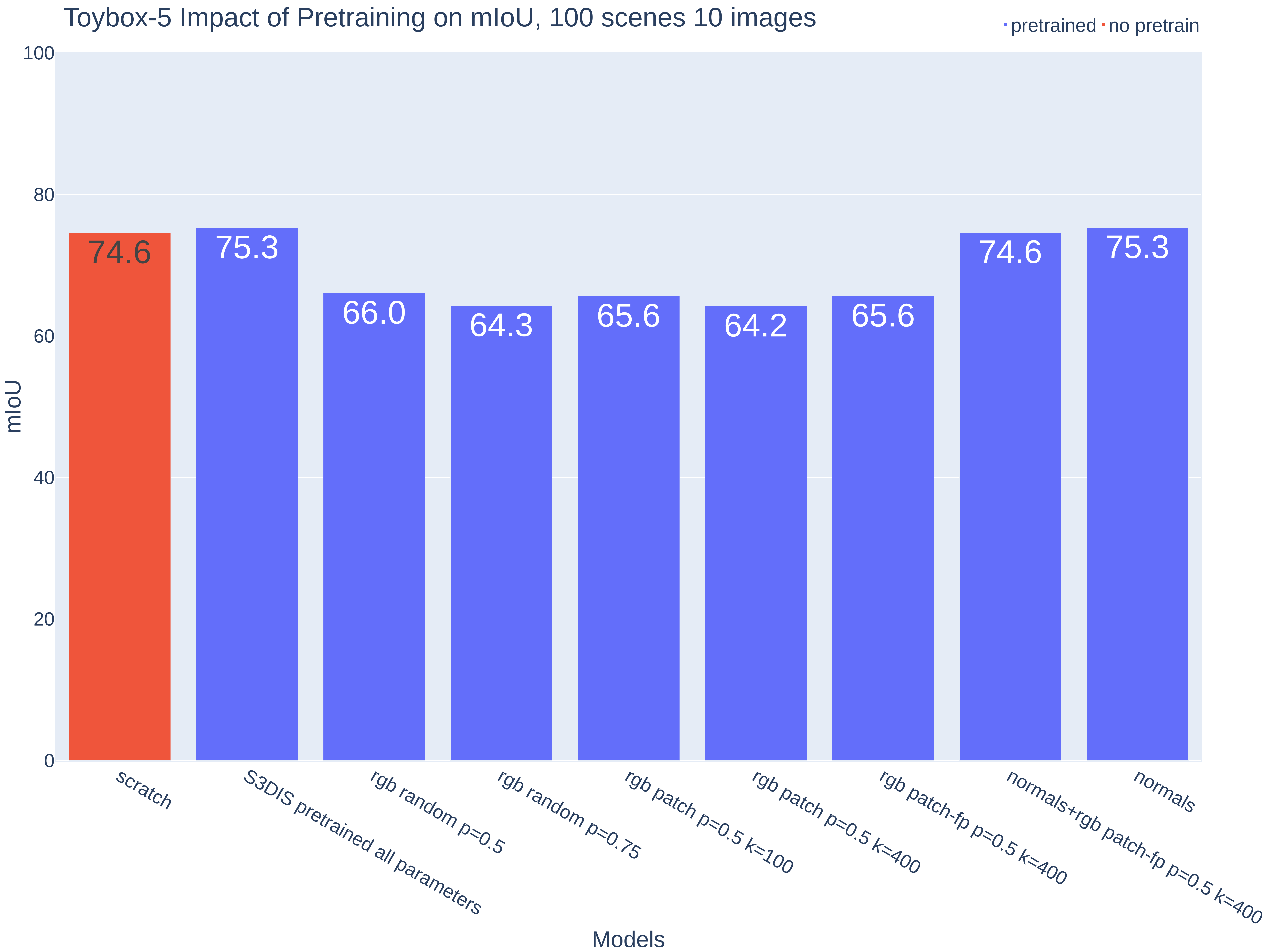}
    \caption{RGB pretraining harms the downstream task's performance significantly. Normal pretraining has the same effect as using S3DIS pretrained weights.}
    \label{fig:toybox_100_10_pretraining}
    \vspace{-1em}
\end{figure}

\begin{figure}[!htbp]
    \centering
    \includegraphics[width=0.5\textwidth]{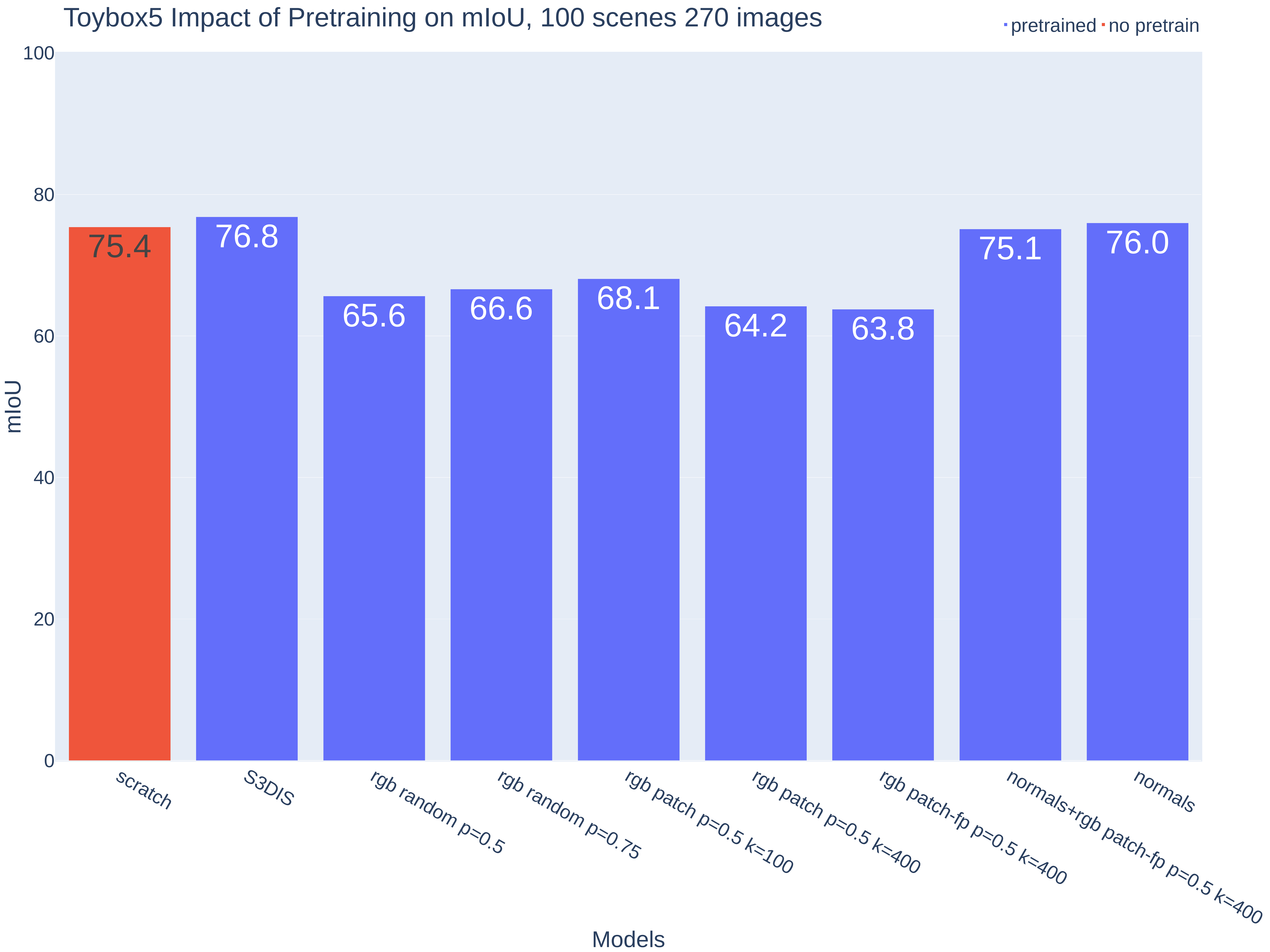}
    \caption{RGB pretraining harms the downstream task's performance significantly. Normal pretraining has the same effect as using S3DIS pretrained weights.}
    \label{fig:toybox_100_270_pretraining}
\end{figure}

\vspace{-3pt}
\section{Conclusion}\label{sec:conclusion}
In this paper, we propose a sample efficient method for 2D novel view and 3D semantic segmentation purely from 2D supervision that accepts any NeRF parameterization by extracting point clouds that capture surfaces in the scene. Caching an intermediate representation, a neural point cloud, speeds up inference time using the proposed Field Head. We show how masked autoencoding and normal pretraining help to boost accuracy in data-scarce scenarios. This work could pave the way for data-driven methods for a wide range of 3D reasoning tasks such as language grounding, scene completion, and navigation.
\newpage
{
    \small
    \bibliographystyle{ieeenat_fullname}
    \bibliography{zotero_references_manual}
}
\clearpage
\setcounter{page}{1}
\maketitlesupplementary

\section{Ablation Studies}
We also conduct ablation studies to justify our design decisions. In the following, we report the impact of proximity loss, ground removal, and surface sampling. Additionally, we show how pretraining normals over big set of scenes improves their accuracies.

\subsection{Point Segmentation Model Choice}
\begin{figure}
    \centering
    \includegraphics[width=0.5\textwidth]{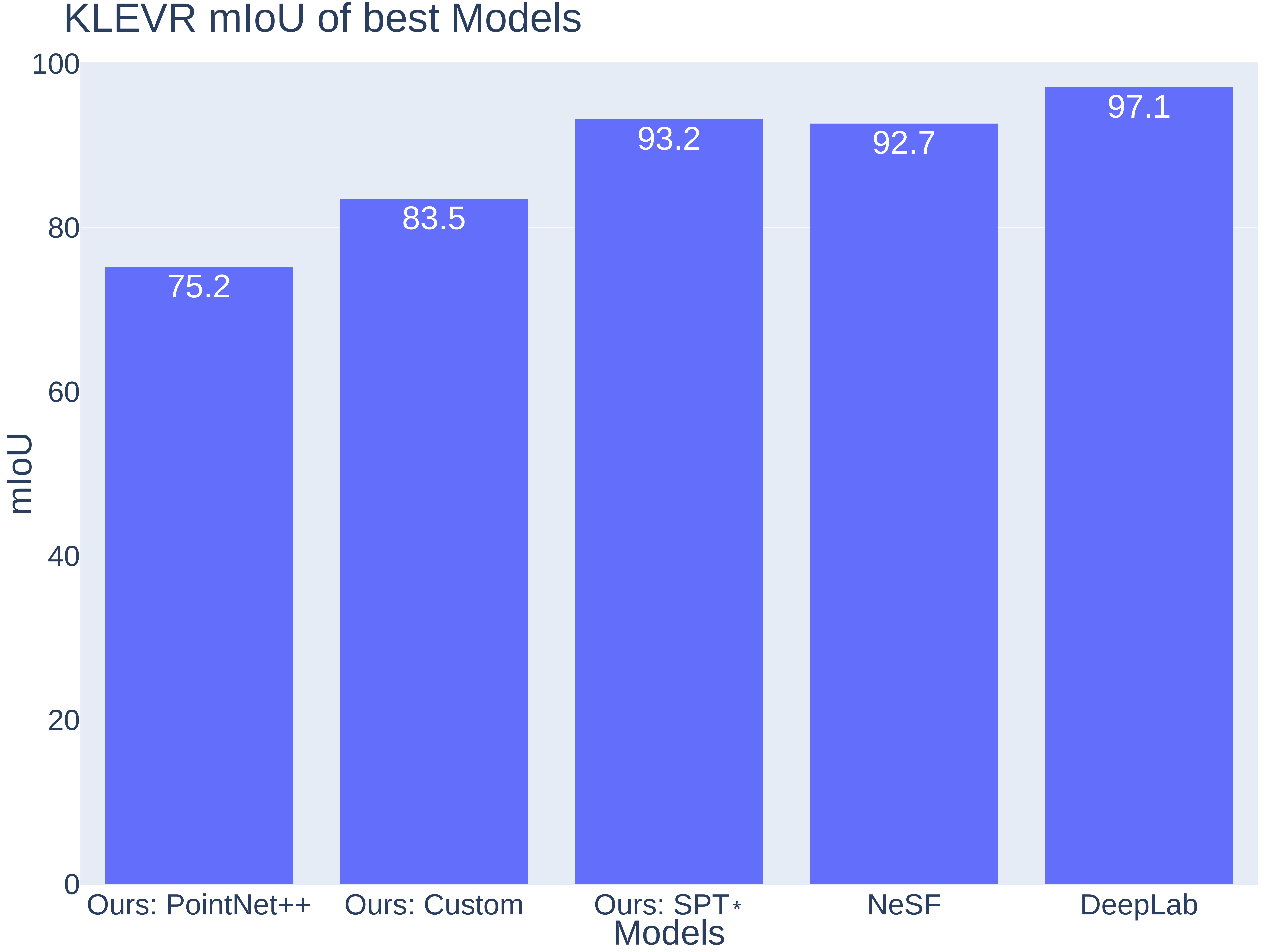}
    \caption{The SPT outperforms all other methods on KLEVR. PointNet++ and Custom are significantly inferior. * indicates the use of S3DIS pretrained weights.}
    \label{fig:klevr_best_trans}
\end{figure}

\begin{figure}
    \centering
    \includegraphics[width=0.5\textwidth]{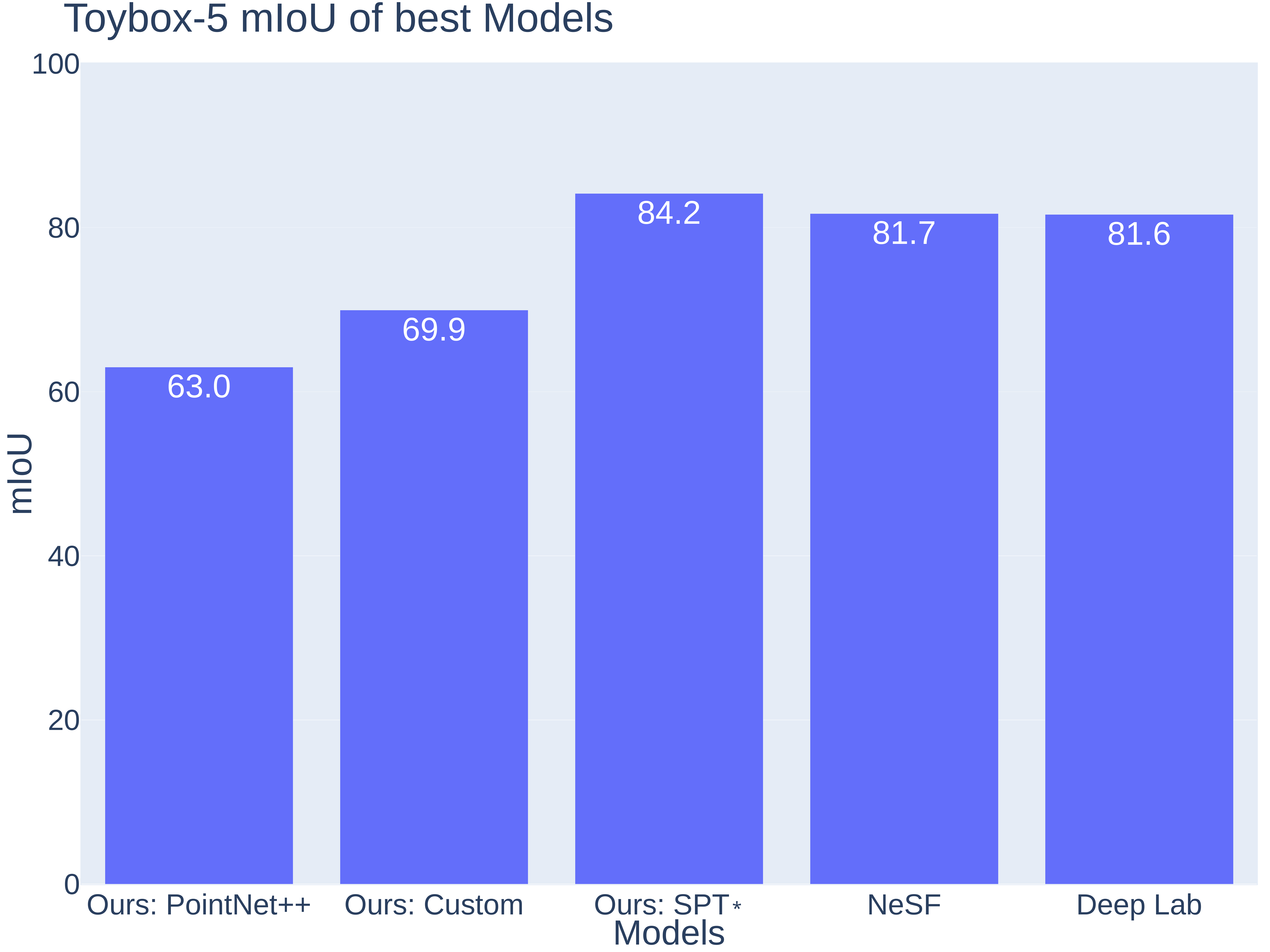}
    \caption{The SPT outperforms all other methods on ToyBox5. PointNet++ and Custom are significantly inferior. * indicates the use of S3DIS pretrained weights.}
    \label{fig:toybox_best_trans}
\end{figure}

The Transformation Network can be any point cloud segmentation network. We compared PointNet++ \cite{qi_pointnet_2017}, a plain custom transformer, and the stratified point transformer \cite{lai_stratified_2022}. On KLEVR and ToyBox5 the modern SPT \cite{lai_stratified_2022} architecture outperformed the naive custom transformer and PointNet++. This is directly noticeable in the qualitative results, see \cref{fig:qualitative}, where PointNet++ and the custom transformer suffer from a lot of uncertainty within objects and also have many classifications of whole objects. In \cref{fig:klevr_best_trans} and \cref{fig:toybox_best_trans} one can quantitatively see how SPT outperforms both other models by roughly 20 mIoU. The custom transformer architecture already outperforms the old PointNet++ architecture. 

\subsection{Ground Removal}
\begin{figure}
    \centering
    \includegraphics[width=0.5\textwidth]{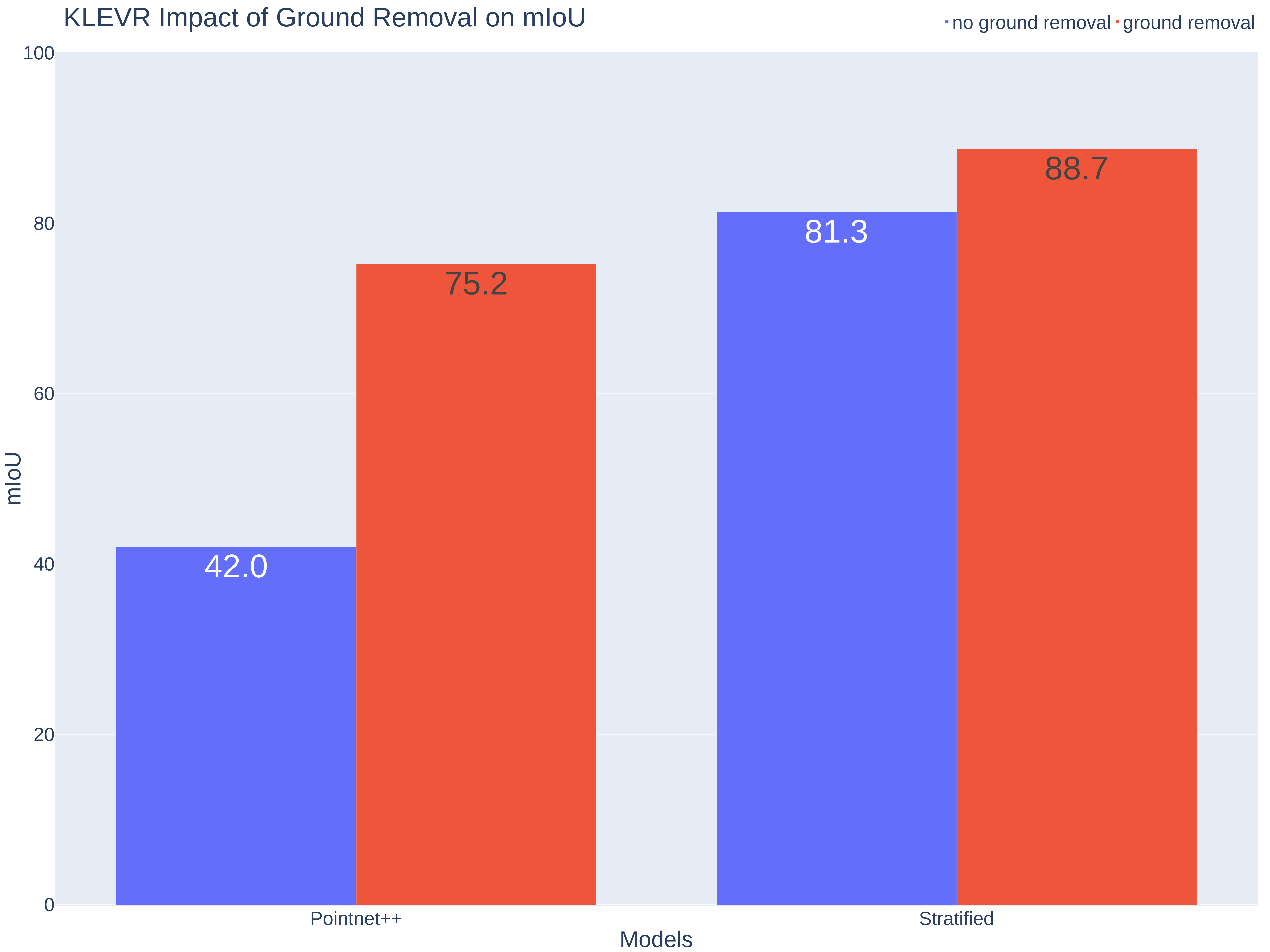}
    \caption{The impact of ground for KLEVR.}
    \label{fig:klevr_ground_removal}
\end{figure}

We show that our ground removal is not only helpful for accuracy but also accelerating inference. To this end, we trained PointNet++ and the SPT on KLEVR once and without ground removal. The custom architecture is too memory consuming hence we skip the evaluation. In \cref{fig:klevr_ground_removal} one can see clearly how both methods achieve higher mIoU when ground removal is used. The impact is bigger on PointNet++ with 32.5 mIoU compared to 7.4 mIoU for the SPT. Additionally, the point clouds are on average 54\% smaller in size. Therefore, we benefit from faster processing time as well. Given the results, we decided to always use Ground Removal for the main experiments.

\subsection{Surface sampling}
\begin{figure}
    \centering
    \includegraphics[width=0.5\textwidth]{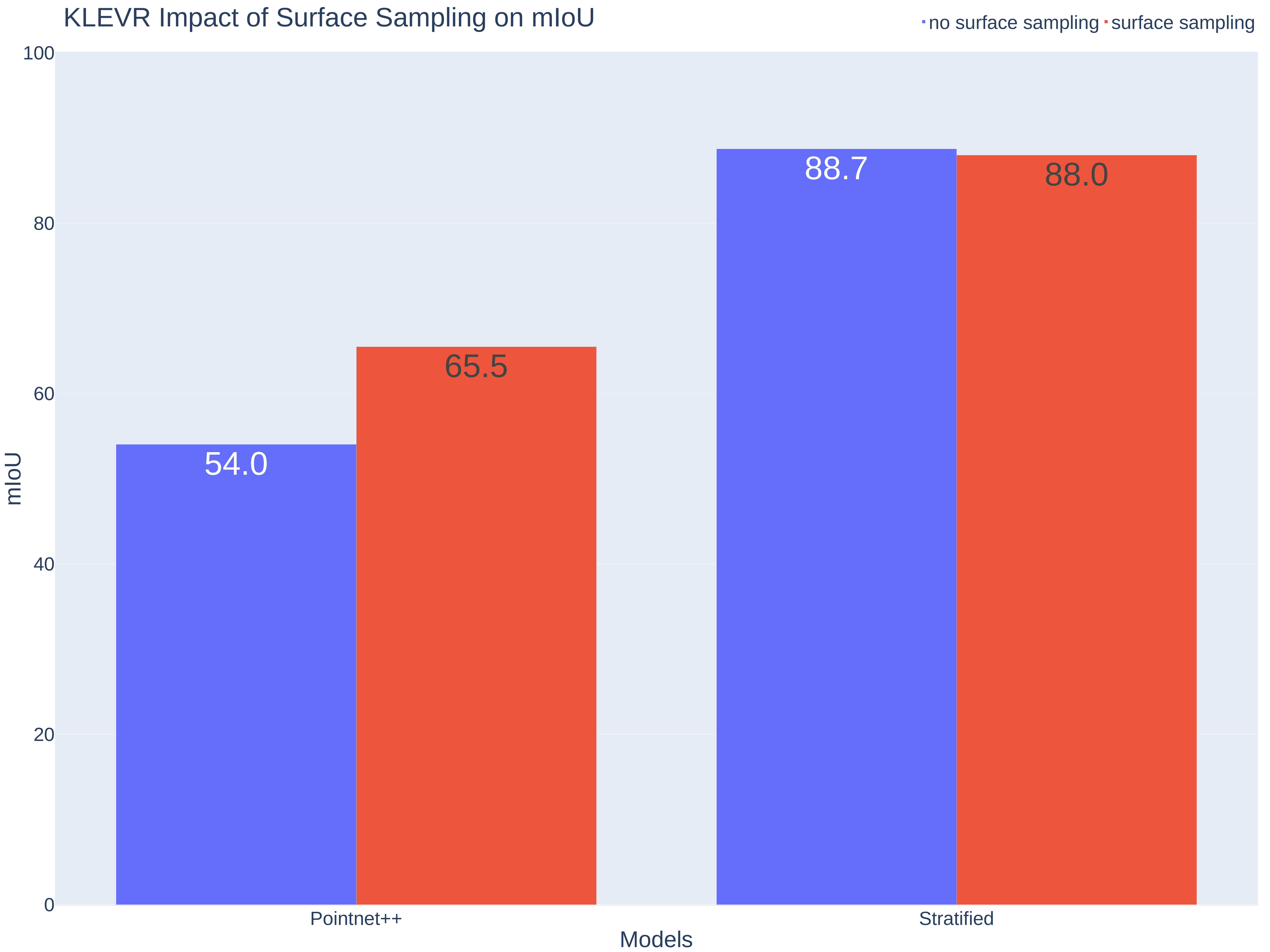}
    \caption{The impact of surface sampling for KLEVR.}
    \label{fig:klevr_surface_sampling}
\end{figure}

Identically to ground removal evaluation we investigate the impact of surface sampling, see \cref{fig:klevr_surface_sampling}. Here, the benefits are smaller in terms of accuracy gain, but both PointNet++ and the SPT still yield higher mIoU by 11.5/0.7 with surface sampling enabled. Due to the significantly higher number of samples per ray, we decreased the total rays per batch from $65.536$ to $8.912$ with 8 samples per ray to yield a similarly big point cloud. Thus, surface sampling is not only beneficial for accuracy but also allows the processing of significantly larger batch sizes/higher amounts of rays. Surface sampling is used for all other experiments. 

\subsection{Proximity Loss}
\begin{figure}
    \centering
    \includegraphics[width=0.5\textwidth]{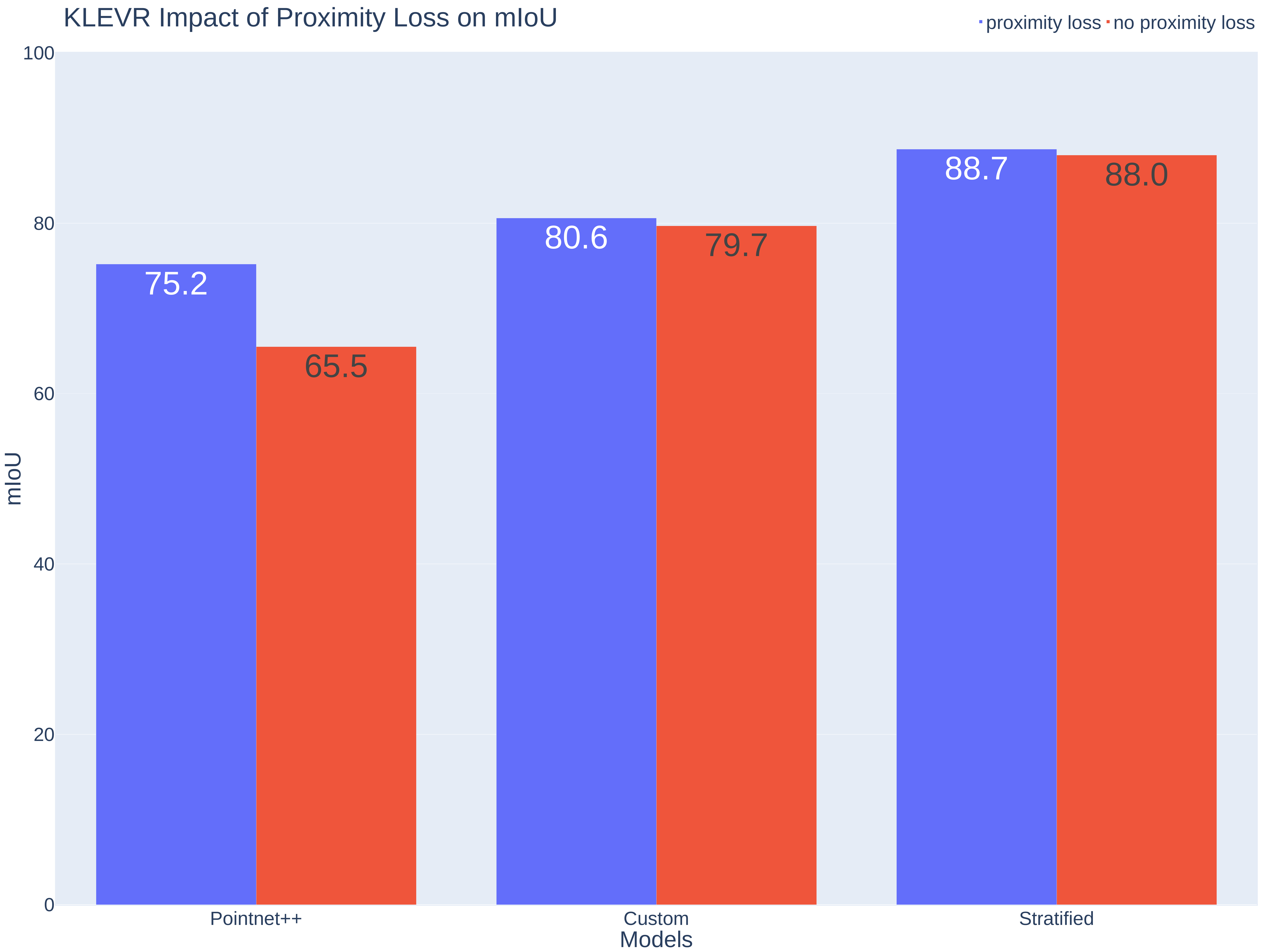}
    \caption{The impact of proximity loss for KLEVR.}
    \label{fig:proximity loss}
\end{figure}

The proximity loss is supposed to encourage locally smooth predictions. We show that for all transformation models, it increases the performance. We trained all 3 transformation baseline models with and without the proximity loss. We can observe the largest effect for PointNet++ but even the custom transformer and the stratified point transformer benefit from the proximity loss, see \cref{fig:proximity loss}. While the proximity loss has the downside of slower training time, because each point cloud needs to be transformed twice, we use the proximity loss for all other experiments due to the observed accuracy gains. 

\subsection{Normal Pretraining}
\begin{figure}
    \centering
    \includegraphics[width=0.5\textwidth]{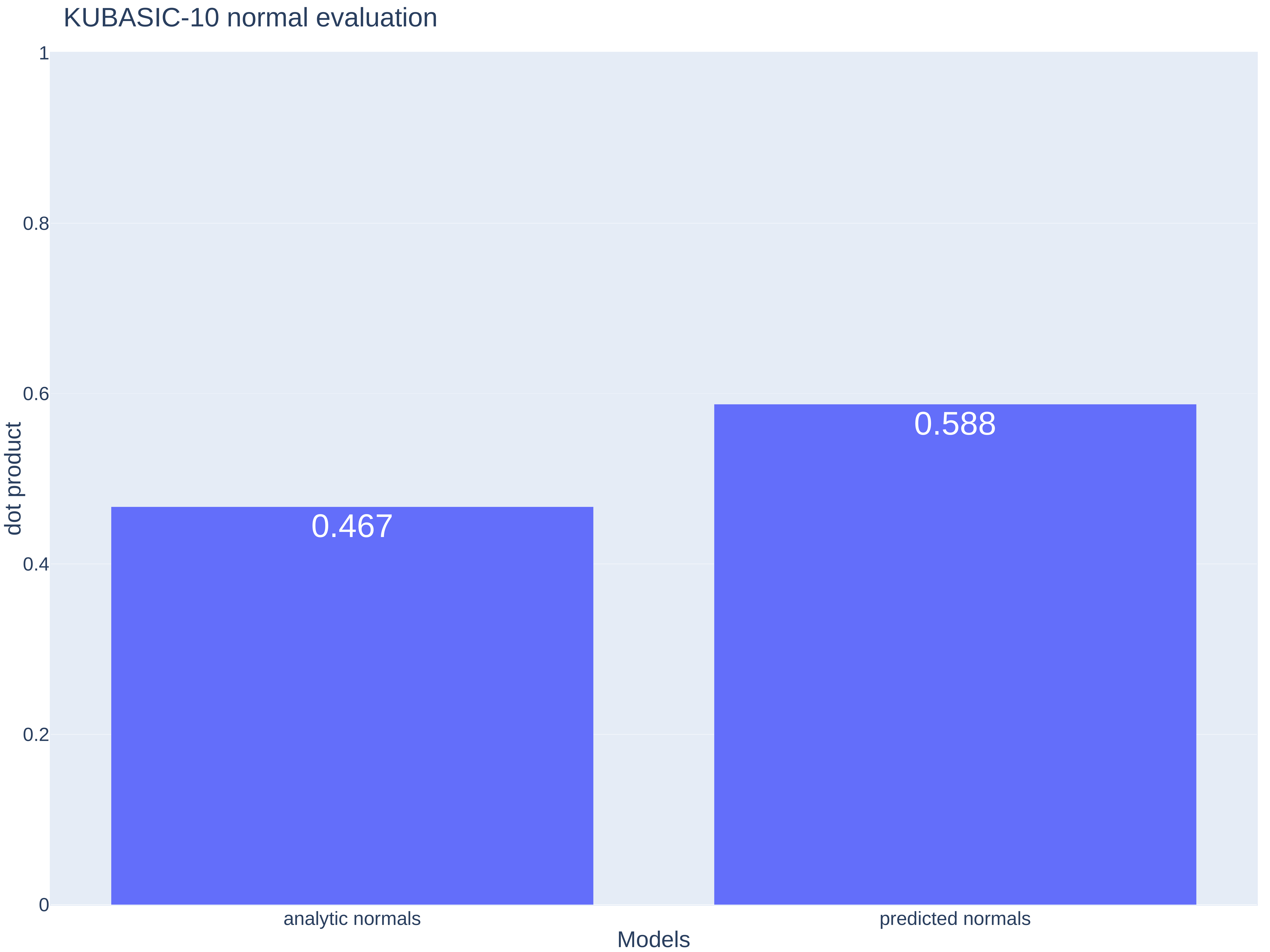}
    \caption{The learned normals during the pertaining phase are better than the analytic normals of the NeRF. Here we report the mean dot product in the eval set of KUBASIC}
    \label{fig:pretrain_normals}
\end{figure}

\begin{figure*}[h]
    \centering
    \begin{subfigure}[b]{0.8\textwidth}
        \centering
        \includegraphics[width=\textwidth]{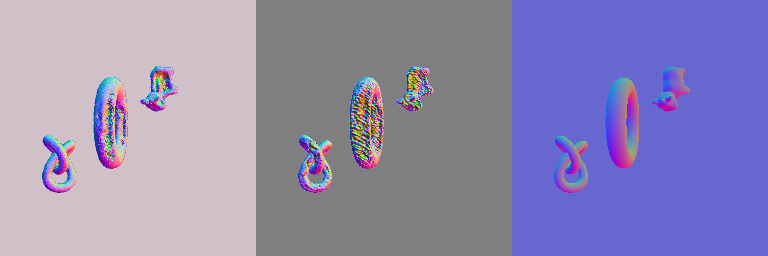}
    \end{subfigure} \\
    \begin{subfigure}[b]{0.8\textwidth}
        \centering
        \includegraphics[width=\textwidth]{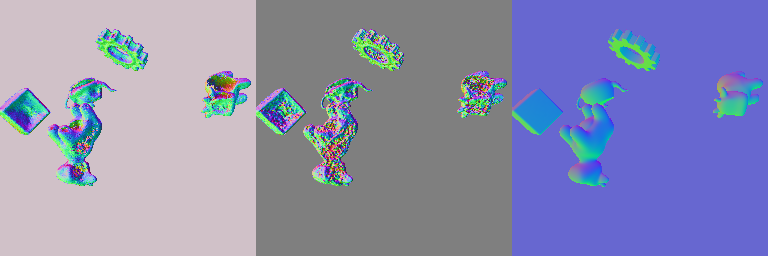}
    \end{subfigure} \\
    \begin{subfigure}[b]{0.8\textwidth}
        \centering
        \includegraphics[width=\textwidth]{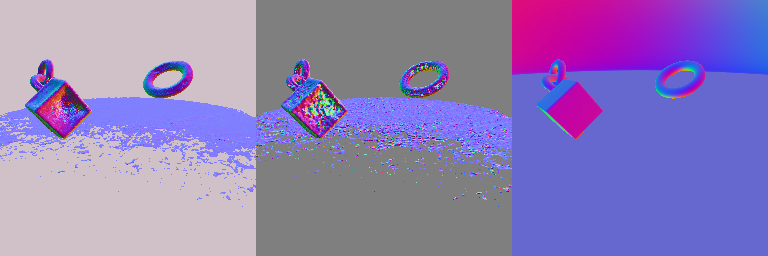}
    \end{subfigure} \\
    \begin{subfigure}[b]{0.8\textwidth}
        \centering
        \includegraphics[width=\textwidth]{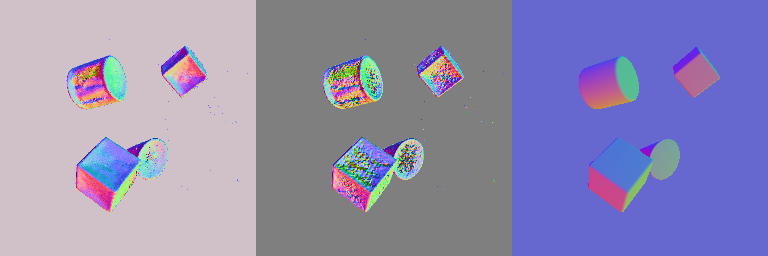}
    \end{subfigure} \\
    \caption{Normal Evaluation on Kubasic. From left to right, normals of the transformation model during the pretrained stage, the supervising normals provided by the NeRF, and the ground truth normals purely used for evaluation. The normals obtained by the transformation model during the pretraining stage are significantly smoother and more accurate than the ones it learned on.}
    \label{fig:normal_qualitative}
\end{figure*}

During the pretraining stage for learning normals, we noticed remarkably smoother predictions compared to NeRF. To validate the pre-trained model, we test it on our custom KUBASIC dataset, which includes ground truth normals. The normals learned by $\mathcal{T}$ outperform the NeRF-provided normals by 25\% in dot product alignment (\cref{fig:pretrain_normals}). Qualitatively, the NeRF-provided normals exhibit noise and inaccuracies, while the pretraining-learned normals are smoother and more precise (\cref{fig:normal_qualitative}). We attribute this to the abundance of scenes available during pretraining, allowing $\mathcal{T}$ to achieve a generally good estimate of normals by transferring knowledge from other scenes and objects, even though it might not achieve a perfect fit for each noisy, individual scene. Learning on a larger corpus of scenes to enhance NeRF normals and fidelity presents an avenue outlook for future research.

\subsection{Field Head}
It is not required that the field-to-field head uses a surface sampled point cloud nor is it required that the query point cloud is surface sampled. We evaluate all combinations on both KLEVR and ToyBox5, see \cref{tab:klevr_field_head} and \cref{tab:toybox_field_head}. On KLEVR all combinations provide the same results. We suspect that its simplicity is not enough to highlight differences. On ToyBox5 we can observe that surface-sampled query point clouds are beneficial, as proposal-based query point clouds lead to fuzzy and unsharp edges. Additionally, fine details get lost more frequently. It is most likely required to significantly increase the density of the neural point cloud to benefit from a proposal-based query. However, this brings up the computational cost significantly. A proposal-based neural point cloud does not show any benefits, we suspect that as the first surface point carries the most sway during volumetric rendering (\cref{eq:volume_rendering}) the additional points do not contribute helpful additional value. 
\begin{table}[]
    \centering
    \begin{tabular}{c|c|c}
        Query Point Cloud & Neural Point Cloud & mIoU \\
        \hline
        Surface (32,768) & Surface (65,536) & 90.0 \\
        Surface (32,768) & Proposal (24,576) & 89.7 \\
        Proposal (8,192) & Surface (65,536) & 90.0\\
        Proposal (8,192) & Proposal (24,576) & 90.1 \\
    \end{tabular}
    \caption{Stratified point transformer with the field-to-field head in different versions of sampling point clouds on KLEVR. In brackets the number of rays used for the sampling method.}
    \label{tab:klevr_field_head}
\end{table}
\begin{table}[]
    \centering
    \begin{tabular}{c|c|c}
        Query Point Cloud & Neural Point Cloud & mIoU \\
        \hline
        Surface (32,768) & Surface (65,536) & 81.9 \\
        Surface (32,768) & Proposal (24,576) & 81.5 \\
        Proposal (4,096) & Surface (65,536) & 75.3 \\
        Proposal (8,192) & Proposal (24,576) & 75.3 \\
    \end{tabular}
    \caption{Stratified point transformer with the field-to-field head in different versions of sampling point clouds on ToyBox5.}
    \label{tab:toybox_field_head}
\end{table}

\section{Confusion Matrices}
We report the normalized confusion matrices of our methods on ToyBox5.

\begin{figure*}
    \centering
    \includegraphics[width=\textwidth]{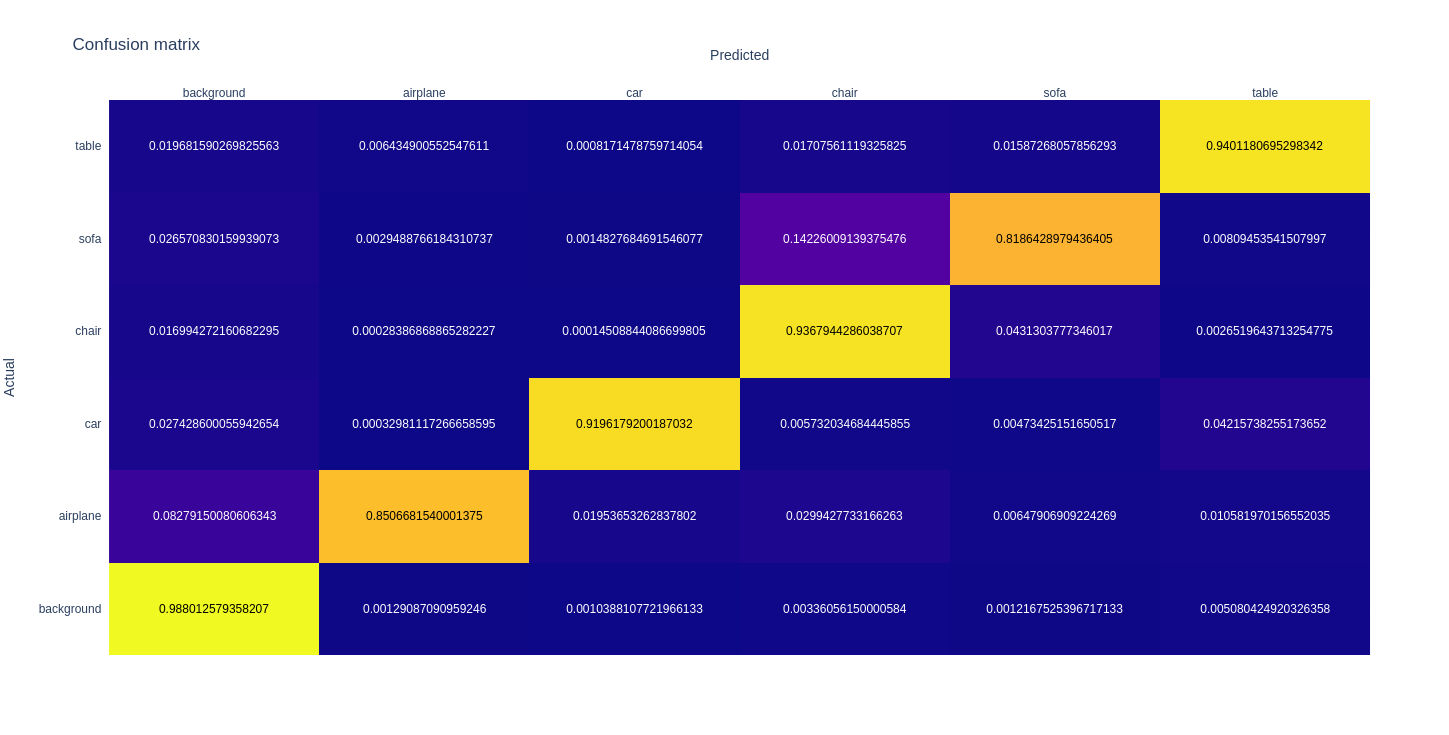}
    \caption{Confusion matrix for the SPT (S3DIS pretrained weights) on ToyBox5.}
    \label{fig:confusion_spt}
\end{figure*}

\begin{figure*}
    \centering
    \includegraphics[width=\textwidth]{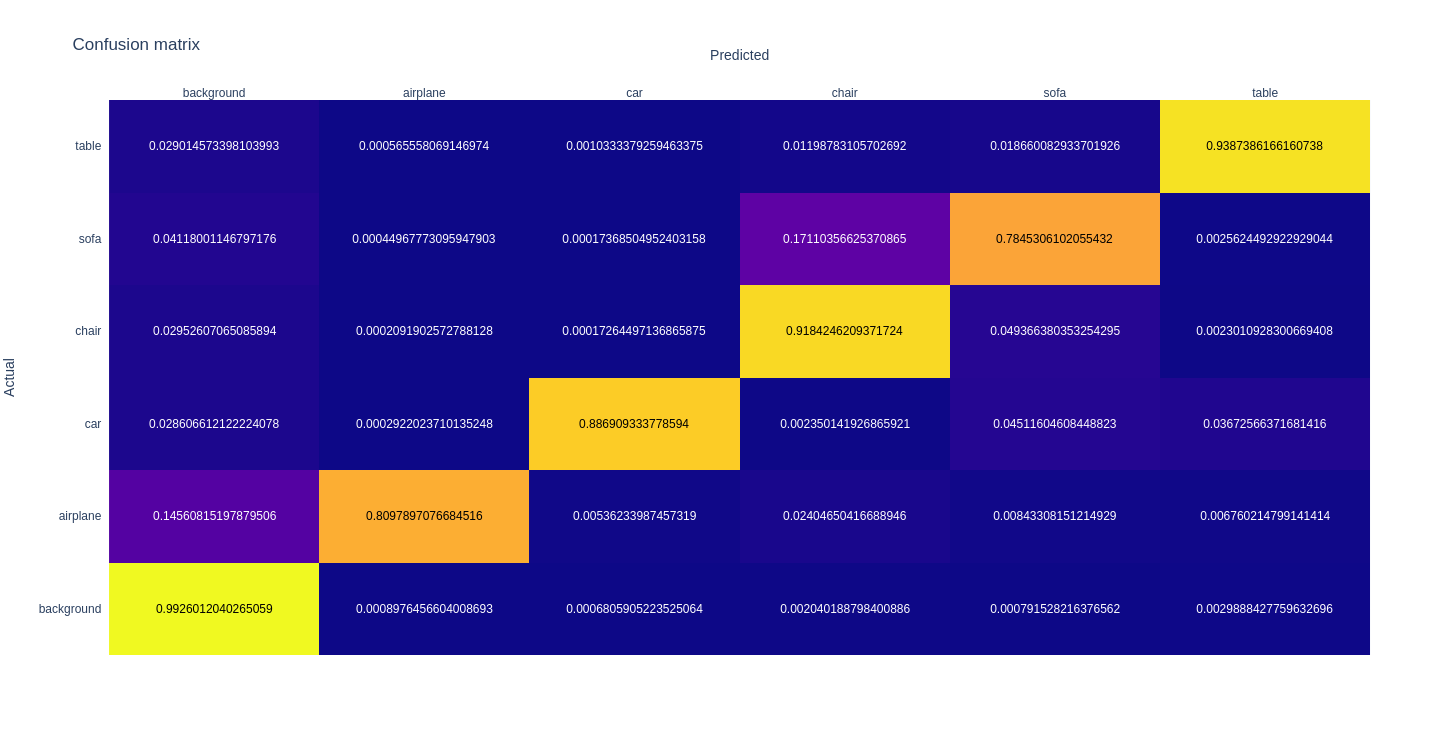}
    \caption{Confusion matrix for the SPT (S3DIS pretrained weights) + FieldHead on ToyBox5.}
    \label{fig:confusion_field2field}
\end{figure*}

\begin{figure*}
    \centering
    \includegraphics[width=\textwidth]{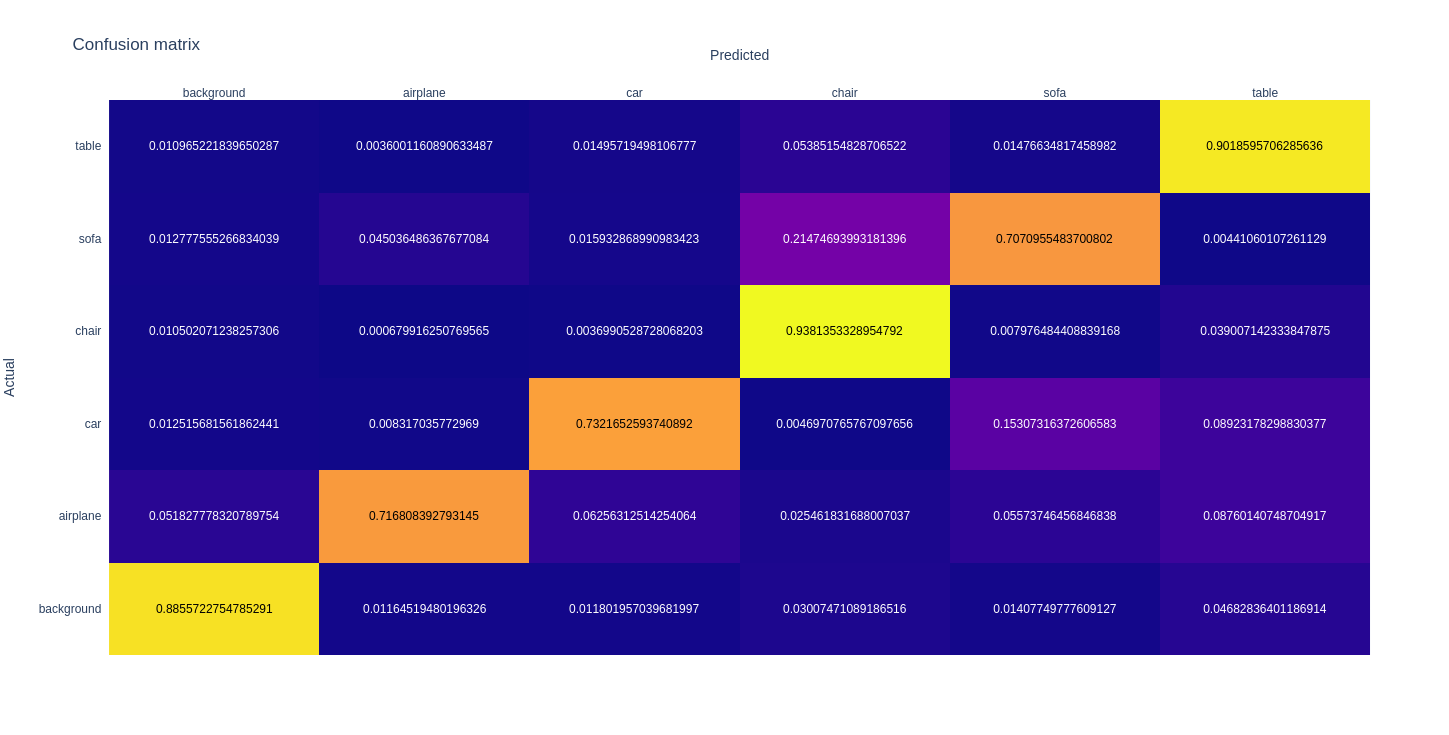}
    \caption{Confusion matrix for the custom transformer on ToyBox5.}
    \label{fig:confusion_custom}
\end{figure*}

\begin{figure*}
    \centering
    \includegraphics[width=\textwidth]{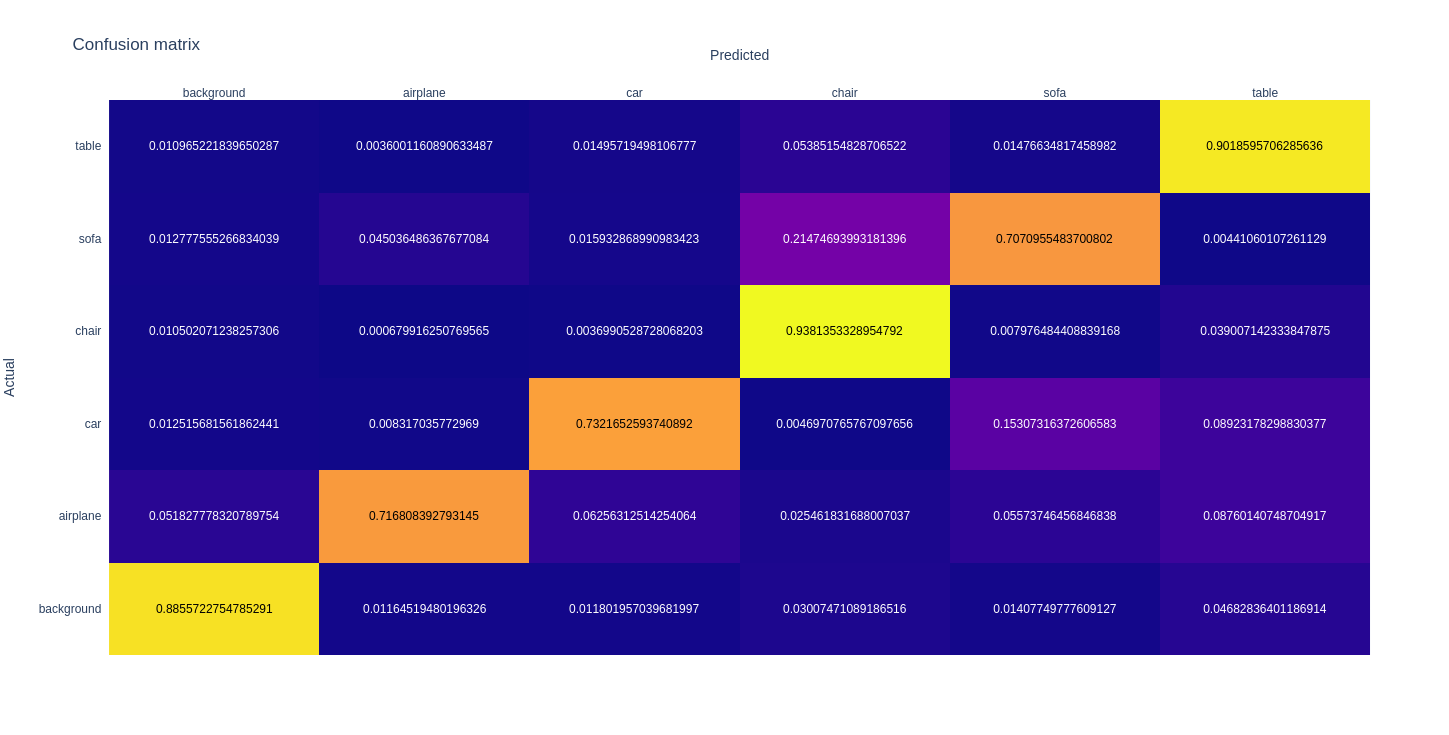}
    \caption{Confusion matrix for the custom transformer on ToyBox5.}
    \label{fig:confusion_custom}
\end{figure*}

\end{document}